\documentclass[12pt]{article}

\usepackage{epicpaper}
\usepackage{times}

\usepackage[utf8]{inputenc}
\usepackage[T1]{fontenc}

\usepackage{amsmath,amssymb,amsthm}
\usepackage{mathtools}
\usepackage{nicefrac}
\usepackage{microtype}
\usepackage{booktabs}
\usepackage{array}      
\usepackage{algorithm}
\usepackage{algpseudocode}
\usepackage{graphicx}
\usepackage{placeins}   
\usepackage{multirow}
\usepackage{hyperref}
\usepackage{cleveref}
\usepackage{xcolor}
\usepackage{enumitem}
\usepackage{wrapfig}
\usepackage{booktabs}
\usepackage{xcolor}
\usepackage{pifont}

\definecolor{stabgreen}{RGB}{120, 190, 130}
\definecolor{staborange}{RGB}{240, 165, 60}
\definecolor{stabred}{RGB}{210, 70, 70}

\newcommand{\stabok}{{\color{stabgreen}\ding{108}}}
\newcommand{\stabwarn}{{\color{staborange}\ding{108}}}
\newcommand{\stabbad}{{\color{stabred}\ding{108}}}
\newcommand{\QuadraSHAP}{\textsc{QuadraSHAP}}

\newcommand{\timeout}{\textcolor{red}{\textsc{t/o}}}

\crefname{equation}{equation}{equations}
\Crefname{equation}{Equation}{Equations}

\crefname{equation}{equation}{equations}
\Crefname{equation}{Equation}{Equations}

\newtheorem{proposition}{Proposition}
\newtheorem{lemma}{Lemma}
\newtheorem{definition}{Definition}
\newcommand{\cD}{\mathcal{D}}
\newcommand{\cS}{\mathcal{S}}
\newcommand{\cT}{\mathcal{T}}
\newcommand{\cE}{\mathcal{E}}
\newcommand{\cL}{\mathcal{L}}

\newcommand{\E}{\mathbb{E}}
\newcommand{\R}{\mathbb{R}}
\newcommand{\X}{\mathbf{X}}
\newcommand{\x}{\mathbf{x}}

\newcommand{\pathD}{\eta}

\usepackage{soul}

\graphicspath{{./figures/}}

\title{QuadraSHAP: $\epsilon$-Exact Shapley Values for Product Games in Logarithmic Parallel Time}

\author{
    Majid Mohammadi\textsuperscript{\rm 1} \and 
    Grigory Reznikov\textsuperscript{\rm 2,3} \and 
    Pavel Sinitcyn\textsuperscript{\rm 2,3} \and 
    Krikamol Muandet\textsuperscript{\rm 4} \and 
    Siu Lun Chau\textsuperscript{\rm 5} \and 
}
\vspace{-1em}
\date{
\footnotesize{
    \textsuperscript{\rm 1} Department of Biomedical Data Science, Leiden University Medical Center, The Netherlands \\
    \textsuperscript{\rm 2} AI Technology for Life, Information and Computing Sciences, Utrecht University, The Netherlands \\
    \textsuperscript{\rm 3} Biomolecular Mass Spectrometry and Proteomics, Pharmaceutical Sciences, Utrecht University, The Netherlands \\    
    \textsuperscript{\rm 4}Rational Intelligence Lab, CISPA Helmholtz Center for Information Security, Germany\\
    \textsuperscript{\rm 5}Epistemic Intelligence \& Computation Lab, College of Computing \& Data Science, Nanyang Technological University, Singapore\\[2ex]}
    {\normalsize \today}
}

\begin{document}
\maketitle

\begin{abstract}
We introduce \textsc{QuadraSHAP}, a method for \emph{$\epsilon$-exact} Shapley computation in \emph{product games}, cooperative games whose coalition values factorize across players. Given a tolerance $\epsilon\ge0$, the method determines a sufficient computational budget before evaluation, guaranteeing an absolute attribution error of at most $\epsilon$ for every feature in exact arithmetic. By extending to weighted sums of product games, the framework supports baseline and empirical interventional attribution across a broad class of models, including log-link regression models, Cox proportional-hazards models, odds-scale classifiers, product-kernel machines, and tree-based models. For a $d$-player product game, we replace the exponentially large coalition sum by a one-dimensional integral of a polynomial of degree at most $d-1$. Gauss--Legendre quadrature with $m_q=\lceil d/2\rceil$ nodes therefore recovers exact Shapley values when $\epsilon=0$. For positive tolerances, a computable error bound derived from the game factors decays geometrically with the node count and can certify substantially smaller budgets. To support efficient evaluation at scale, we share computations across features and evaluate products in log-space with sign tracking, mitigating intermediate overflow and underflow. Given the quadrature rule, computing all $d$ attributions requires $O(d\,m_q)$ work per product game, compared with $O(2^d)$ for direct coalition summation, and admits $O(\log d)$ parallel time with sufficient processors. On a survival analysis problem with $396{,}065$ features, the exact quadrature configuration computes all feature attributions in approximately five minutes of GPU evaluation per explanation, at a scale where exhaustive coalition enumeration is infeasible. Explicit error control enables a further reduction: setting $\epsilon=10^{-6}$ selects a smaller, theoretically certified quadrature budget and reduces the mean evaluation time to $0.36$ seconds per explanation. These results demonstrate scalable attribution with explicit error control and motivate application of low-latency explanations for real-time safety-critical decision support. Code is available at \url{https://github.com/majeed7/QuadraSHAP}.

\end{abstract}

\newpage 
\stoptoc
\section{Introduction}

Shapley values (SVs)~\citep{shapley} explain model predictions by assigning feature contributions according to axioms from cooperative game theory~\citep{shap}. Their definition averages each feature's contribution over subsets of features, called \emph{coalitions}. Direct evaluation involves exponentially many coalitions, making exact computation expensive in general. Existing methods address this cost through model-agnostic approximation~\citep{shap,shapiq,polyshap} or algorithms that exploit the structure of particular model classes~\citep{treeshap,linear_treeshap,rkhs_shap,gpshap,fgpx_shapley}.

We focus on a shared structure that makes a common computation possible: coalition values that factorize across features. In a \emph{product game}~\citep{cooperative_prodcut_games}, the value assigned to a coalition $\cS$ takes the form $$v(\cS)=\prod_{j\in\cS}u_j,$$ where $u_j$ is the factor associated with feature $j$ (\Cref{def:product_game}). By linearity, the SVs of weighted sums of these games follow from the same computation. Extending this formulation to products with factors for both present and absent features covers log-link generalized linear and additive models, Cox proportional-hazards models, logistic regression and naive Bayes on the odds scale, product-kernel machines, and rule ensembles, including decision trees (\Cref{tab:model_classes}).

Exploiting this structure requires both computational efficiency and numerical reliability. Polynomial interpolation can eliminate coalition enumeration, but recovering high-degree polynomial coefficients can amplify numerical errors~\citep{linear_treeshap}. Independently, directly multiplying thousands of feature-wise factors can produce values outside the range of floating-point arithmetic. At these dimensions, even polynomial-time exact evaluation can be costly. A useful method should therefore support a controlled trade-off between accuracy and computation while addressing the numerical difficulties of long products.

We introduce \textsc{QuadraSHAP}, which selects a computational budget to meet a user-specified accuracy target. For a specified game and tolerance $\epsilon\ge0$, we call a computation \emph{$\epsilon$-exact} if the absolute error in every feature's Shapley value is at most $\epsilon$ in exact arithmetic. Setting $\epsilon=0$ recovers exact Shapley values, while positive tolerances can permit smaller computational budgets. The guarantee controls quadrature error: the error introduced by approximating an integral with a weighted sum of function evaluations. Error guarantees are also available for other Shapley estimators: sampling-based methods can provide finite-sample probabilistic guarantees~\citep{maleki2013bounding}, and quadrature over permutations admits deterministic error bounds~\citep{mitchell2022sampling}. Our contribution is to exploit product structure to derive an explicit relationship between accuracy and computational budget. Whereas multilinear sampling methods estimate coalition averages within an integral representation~\citep{okhrati2021multilinear}, we compute these averages analytically, reducing the problem to a one-dimensional polynomial integral. From the factors of this polynomial, we derive a computable quadrature-error bound that determines a sufficient node count before evaluation.

While this work was under development, TreeGrad-Shap~\citep{treegrad_ranker} and Quadrature-TreeSHAP~\citep{quadrature_treeshap} appeared, both using Gauss--Legendre (GL) quadrature for path-dependent tree attribution. For a given node count, our tree specialization evaluates the same leaf-wise integrand at the same GL nodes. We develop this computation for the broader class of product games, together with a computable accuracy guarantee and an evaluator designed for long products. \Cref{app:related_quadrature} provides a detailed comparison.

\textbf{Our contributions are threefold:} 

\paragraph{First, we provide $\epsilon$-exact computation with a sufficient budget determined before evaluation.}
By expressing the Shapley weights as integrals, we replace the exponential coalition sum for a $d$-player product game by a one-dimensional integral of a polynomial of degree at most $d-1$ (\Cref{sec:product_games_glq}). GL quadrature with $m_q=\lceil d/2\rceil$ nodes gives exact computation. For positive tolerances, we derive a geometrically decaying error bound that is computable from the game's factors and can certify smaller node counts (\Cref{prop:node_budget} and \Cref{sec:5:approx}). The exactness threshold remains a fallback whenever no smaller budget is certified.

\paragraph{Second, we share computation across features through log-space evaluation and parallel reductions.}
At each quadrature node, the feature attributions depend on products that differ by only one factor. Reusing these products gives $O(d\,m_q)$ work per game, while accumulating logarithms and tracking signs mitigates intermediate overflow and underflow (\Cref{subsec:stable_efficient}). Combining partial sums and products in a binary tree yields $O(\log d)$ parallel time for $m_q=O(d)$, given the quadrature rule and sufficient processors (\Cref{prop:complexity}). At the exactness threshold $m_q=\lceil d/2\rceil$, this evaluator uses $O(d^2)$ total work.

\paragraph{Third, we give a common formulation across models and attribution settings.}
For models expressed as weighted sums of feature-wise products, we derive the factors needed for baseline attribution, which replaces absent features with values from one reference input, and empirical interventional attribution, which averages these replacements over a background dataset. The same formulation covers neutral-factor attribution, which replaces absent multiplicative factors by one. Each setting reduces to weighted product evaluations, so the same algorithm and error analysis apply once the factors are constructed (\Cref{sec:multiplicative_models}).

\section{Preliminaries}
\label{sec:preliminaries}

\paragraph{Notation.}
Let $\cD=\{1,\ldots,d\}$ index the input features, and let $2^\cD$ denote the collection of all feature subsets, or coalitions. We write $\x\in\mathbb{R}^d$ for the input to be explained, $X$ for a random input, and $\X\in\mathbb{R}^{n\times d}$ for the training feature matrix. For $\cS\subseteq\cD$, $\x_{\cS}$ and $\X_{\cS}$ restrict the vector and matrix to the corresponding coordinates and columns. Expectation is denoted by $\E$. Proofs are deferred to Appendix~\ref{apx:proofs} and the associated technical appendices.

\paragraph{Shapley values.}
To assign feature contributions, a value function $v:2^\cD\to\mathbb{R}$ first specifies a value for each coalition. The Shapley value~\citep{shapley} of feature $i$ averages the change caused by adding $i$ to a coalition, with weights determined by coalition size:
{\small
\begin{equation}
\label{eq:shapley_def}
\phi_i(v)
=
\sum_{\cS\subseteq\cD\setminus\{i\}}
\mu(|\cS|)
\bigl(v(\cS\cup\{i\})-v(\cS)\bigr),
\qquad
\mu(s)=\frac{s!\,(d-s-1)!}{d!}.
\end{equation}
}
Here, $v(\cS\cup\{i\})-v(\cS)$ is the \emph{marginal contribution} of feature $i$, and $\mu(s)$ is its \emph{Shapley weight}. SVs are uniquely characterized by allocating the total difference $v(\cD)-v(\varnothing)$ (\emph{efficiency}), treating interchangeable features equally (\emph{symmetry}), assigning zero to features with no marginal contribution (\emph{null player}), and respecting weighted sums of games (\emph{linearity})~\citep{shapley,fourier_treeshap}. Direct evaluation of \cref{eq:shapley_def} requires $2^{d-1}$ coalition terms per feature, motivating the use of structure in $v$.

\paragraph{Value functions for explainability.}
For a predictive model $f$ and an input $\x$, the value function specifies how to evaluate the model when only the features in $\cS$ are treated as present~\citep{many_sv,shap_book}. One approach averages over absent features conditional on the observed present features, giving the \emph{observational} or \emph{conditional} value
$$
v_{\x}^{\mathrm{obs}}(\cS)
=
\E\bigl[f(X)\mid X_{\cS}=\x_{\cS}\bigr].
$$
Alternatively, drawing the absent features jointly from their marginal distribution gives the \emph{interventional} or \emph{marginal} value
$$
v_{\x}^{\mathrm{int}}(\cS)
=
\E_{X_{\cD\setminus\cS}}
\bigl[f(\x_{\cS},X_{\cD\setminus\cS})\bigr].
$$
This second choice preserves dependence among absent features but does not condition their distribution on the present features~\citep{interventional_shap,chen2020true,causal_shap}.

Our focus is the empirical interventional value function, which replaces this expectation by an average over a fixed background dataset. Using one background input gives \emph{baseline attribution}. For multiplicative models, we also consider \emph{neutral-factor attribution}, which replaces absent multiplicative factors by one (\Cref{app:value_functions}).

\paragraph{Prescribed accuracy: $\epsilon$-exactness.}
To make the accuracy--computation trade-off explicit, we allow the user to prescribe an absolute error tolerance for every feature. For a fixed game $v$ and $\epsilon\ge0$, we call a computation \emph{$\epsilon$-exact} if its outputs satisfy, in exact arithmetic,
\begin{equation}
\label{eq:epsilon_exact}
\max_{i\in\cD}
\bigl|\widehat{\phi}_i(v)-\phi_i(v)\bigr|
\le\epsilon.
\end{equation}
Setting $\epsilon=0$ recovers exact SVs, while positive tolerances permit approximation within the prescribed accuracy. Our guarantees control numerical integration error for the specified game; they do not include floating-point rounding or the (statistical) error from replacing a population expectation by a finite background average.

\paragraph{Product games.}
We begin with games in which each present feature contributes a fixed multiplicative factor, while each absent feature contributes one~\citep{cooperative_prodcut_games}.

\begin{definition}[Product game]
\label{def:product_game}
A value function $v:2^\cD\to\mathbb{R}$ is a \emph{product game} if there exist factors $u_1,...,u_d\in\mathbb{R}$ such that
$$
v(\cS)=\prod_{j\in\cS}u_j,
$$
with $\cS\subseteq\cD$ and the empty-product convention $v(\varnothing)=1$.
\end{definition}

For feature attribution, these factors are obtained by evaluating feature-wise functions at the input $\x$; we write $u_j:=u_j(x_j)$ for brevity. Adding feature $i$ multiplies the coalition value by $u_i$, so its marginal contribution is
$$
v(\cS\cup\{i\})-v(\cS)
=
(u_i-1)\prod_{j\in\cS}u_j.
$$
Consequently, \cref{eq:shapley_def} becomes
\begin{equation}
\label{eq:shapley_weighted_esp_form}
\phi_i(v)
=
(u_i-1)
\sum_{\cS\subseteq\cD\setminus\{i\}}
\mu(|\cS|)\prod_{j\in\cS}u_j.
\end{equation}
This factorization is the starting point for our integral representation. \Cref{sec:multiplicative_models} extends the computation to weighted sums of products with separate factors for present and absent features. For models with a sum-of-products representation (\Cref{def:multiplicative_model}), this extension covers empirical interventional, baseline, and neutral-factor attribution.

\paragraph{Gauss--Legendre (GL) quadrature.}
Our approach replaces the coalition sum by an integral, then evaluates that integral through a weighted sum at selected points, called \emph{quadrature nodes}. On $[0,1]$, the $m_q$-node GL rule~\citep{davis1984methods} takes the form
$$
\int_0^1 g(t)\,dt
\;\approx\;
\sum_{q=1}^{m_q}\omega_q\,g(\tau_q),
$$
where $\tau_q\in(0,1)$ and $\omega_q>0$ are the nodes and weights. They depend only on $m_q$ and can be precomputed and reused~\citep{golub1969calculation}. The key property is polynomial exactness: in exact arithmetic, the rule integrates every polynomial of degree at most $2m_q-1$ exactly~\citep{trefethen2008gauss}. For the product-game integrals, \Cref{sec:product_games_glq} establishes the resulting exactness threshold and derives error bounds that certify smaller node counts when $\epsilon>0$.

\section{Shapley Values for Product Games via GL Quadrature}
\label{sec:product_games_glq}

Our goal is to compute all Shapley values of a product game to a prescribed absolute accuracy $\epsilon\ge0$, without enumerating feature coalitions. We first express the coalition sum as a one-dimensional integral, then use its product structure to select a sufficient quadrature budget. Finally, we share computations across features to evaluate the resulting rule efficiently. 

\subsection{From Coalition Averages to a One-Dimensional Integral}

The obstacle in \cref{eq:shapley_weighted_esp_form} is the weighted sum over $2^{d-1}$ coalitions. Product structure lets us evaluate this average without visiting individual coalitions. To see why, fix $t\in[0,1]$ and include each feature $j\neq i$ independently with probability $t$. An included feature contributes $u_j$, while an excluded feature contributes one, giving the expected factor $(1-t)+tu_j$. Independence makes the expected product equal to the product of these expected factors. Averaging over $t$ then recovers the Shapley weights.

\begin{proposition}[Integral representation]
\label{prop:shapley_integral_identity_product_game}
Let $v(\cS)=\prod_{j\in\cS}u_j$ be a product game. For every $i\in\cD$,
\begin{equation}
\label{eq:shapley_integral_identity_product_game}
\phi_i(v)
=
(u_i-1)\int_0^1
\bigg(\prod_{j\neq i}1-t + tu_j \bigg)\,dt.
\end{equation}
\end{proposition}
The key identity is the integral representation of the Shapley weights,
$$\mu(s)=\int_0^1 t^s(1-t)^{d-s-1}\,dt.$$
Substituting it into \cref{eq:shapley_weighted_esp_form} allows the coalition sum to be expanded as a product inside the integral. This is also the product-game specialization of Owen's multilinear representation~\citep{owen1972multilinear}, used for tree attribution by TreeGrad-Shap~\citep{treegrad_ranker}; \Cref{app:related_quadrature} compares the formulations. Here, the product structure makes the integrand directly evaluable, with no inner coalition sampling.

\subsection{Choosing a Budget for Prescribed Accuracy}

Each integrand in \cref{eq:shapley_integral_identity_product_game} is a polynomial of degree at most $d-1$. Since an $m_q$-node GL rule integrates polynomials of degree at most $2m_q-1$ exactly, $\lceil d/2\rceil$ nodes suffice for exact SVs. The practical question is whether fewer nodes can certify a positive tolerance.

To bound the error of smaller rules, we control the polynomial's magnitude on an ellipse surrounding the integration interval in the complex plane. For $\rho>1$, let $E_\rho$ denote the \emph{Bernstein ellipse} with foci $0$ and $1$ and semi-major axis $(\rho+\rho^{-1})/4$~\citep{trefethen2008gauss}. The parameter $\rho$ controls the ellipse's size; \Cref{def:nb_bernstein} in Appendix~\ref{app:node_budget} gives its precise construction.

\begin{proposition}
\label{prop:gl_shapley_integral}
For a product game $v$, let ${(\tau_q,\omega_q)}_{q=1}^{m_q}$ be GL nodes and weights on $[0,1]$. Define,
%
\begin{equation}
\label{eq:gl_quad_shapley}
\widehat{\phi}_i(v)
:=
(u_i-1)\sum_{q=1}^{m_q}\omega_q
\prod_{j\neq i}\bigl(1-\tau_q+\tau_q u_j\bigr).
\end{equation}
Then:
\begin{enumerate}[nosep, leftmargin=*]
    \item \emph{Exactness.}
    If $m_q\ge\lceil d/2\rceil$, then
    $\widehat{\phi}_i(v)=\phi_i(v)$ for every $i\in\cD$.

    \item \emph{Error bound.}
    For each $i$, let
    $g_i(z):=\prod_{j\neq i}(1-z+zu_j)$ and
    $M_\rho:=\max_{z\in E_\rho}|g_i(z)|$.
    Then,
    \begin{equation}
    \label{eq:gl_error_bound}
    \bigl|\widehat{\phi}_i(v)-\phi_i(v)\bigr|
    \le
    |u_i-1|\,C_\rho M_\rho\,\rho^{-2m_q},
    \qquad
    C_\rho:=\frac{32}{15}\frac{1}{1-\rho^{-2}}, \quad \forall \rho > 1.
    \end{equation}
\end{enumerate}
\end{proposition}

For fixed $\rho$, the upper bound decreases geometrically with $m_q$. However, it still contains the unknown maximum $M_\rho$. The product structure allows us to bound this quantity directly from the factors. What matters is how much the factors vary relative to their magnitudes, together with the absolute scale of the product.

\begin{proposition}
\label{prop:node_budget}
Let
$$
M_j:=\max(1,|u_j|),\quad 
\Pi_i:=\prod_{j\neq i}M_j,
\quad \Lambda_i:=\sum_{j\neq i}\frac{|1-u_j|}{M_j}.
$$
Then 
\begin{equation}
\label{eq:node_budget}
\begin{aligned}
\bigl|\widehat{\phi}_i(v)-\phi_i(v)\bigr|
&\le |u_i-1|\,\Pi_i\,B(m_q,\Lambda_i)
\end{aligned}
\end{equation}
where $$B(m,\Lambda)
:=
\inf_{\rho>1}
C_\rho
\exp\!\left(
\frac{\Lambda}{4}(\rho+\rho^{-1}-2)
\right)\rho^{-2m}.$$ For $m\ge1$ and $\Lambda\ge0$, $B$ is non-increasing in $m$ and
non-decreasing in $\Lambda$. For $\Lambda>0$, its minimizer is
obtained by solving a scalar equation; for $\Lambda=0$,
$B(m,0)=0$.
\end{proposition}

The quantity $\Lambda_i$ sums the changes from the absent factor one to the present factors, each relative to its magnitude. We call it the \emph{total relative variation}. The prefactor $|u_i-1|\Pi_i$ sets the absolute scale of the error bound. Both quantities therefore influence the certified budget. In general, $\Lambda_i\le2(d-1)$; for nonnegative factors, $\Lambda_i\le d-1$. When many factors are close to one, the total relative variation can remain small even in high dimensions, allowing smaller budgets when the scale of the bound also permits them.

\paragraph{Selecting the node count.}
Let $m_{\mathrm{exact}}=\lceil d/2\rceil$.
For $\epsilon=0$, we use $m_{\mathrm{exact}}$.
For $\epsilon>0$, we search for the smallest
$m<m_{\mathrm{exact}}$ satisfying
$$
\max_{i\in\cD}
|u_i-1|\,\Pi_i\,B(m,\Lambda_i)
\le\epsilon.
$$
Monotonicity permits bisection over the candidate node counts. If no smaller count passes, we return $m_{\mathrm{exact}}$, whose guarantee follows from polynomial exactness rather than from the analytic error bound. Thus, every prescribed tolerance has a sufficient budget. This budget is the smallest accepted by the certificate and exactness fallback, not necessarily the smallest that achieves the requested accuracy in practice. \Cref{app:node_budget} gives the derivation and extends the certificate to weighted sums of games; \Cref{alg:nb_budget} summarizes the procedure.

The resulting computation evaluates products at quadrature nodes, without enumerating coalitions or recovering polynomial coefficients. It therefore avoids the coefficient-recovery step of interpolation-based approaches~\citep{linear_treeshap,tn_shap}, which inverts an ill-conditioned Vandermonde system and can amplify numerical errors~\citep{gautschi1975norm}. Each product, however, has $d-1$ factors, and a product of thousands of factors can overflow or underflow floating-point arithmetic even when every factor is moderate; forming it separately for each feature would also cost $O(d^2 m_q)$ work. The next subsection addresses both: it evaluates the products in log-space so that their magnitude never leaves the representable range, and shares them across features so that the work is $O(d\,m_q)$.


\subsection{Shared Evaluation and Parallel Complexity}
\label{subsec:stable_efficient}

\paragraph{Reusing products across features.}
At each quadrature node, the products required by different features differ by only one factor. Computing each independently would take $O(d^2m_q)$ work. Sharing their common factors reduces this to $O(d\,m_q)$, while log-space arithmetic mitigates the intermediate overflow and underflow that can arise when multiplying many factors.

For each node, define
$$
T_{q,j}:=(1-\tau_q)+\tau_q u_j,
\,\quad 
P_q:=\prod_{j=1}^{d}T_{q,j}.
$$
When every $T_{q,j}$ is nonzero, the product needed for feature $i$
is $P_q/T_{q,i}$. Rather than forming $P_q$ directly, we accumulate its log-magnitude
and track its sign by setting $$\log|P_q|=\sum_{j=1}^{d}\log|T_{q,j}|$$ and $$\mathrm{sgn}(P_q)=\prod_{j=1}^{d}\mathrm{sgn}(T_{q,j}).$$
Removing the corresponding log-magnitude and sign gives
{
\begin{equation}
\label{eq:gl_logspace_phi}
\widehat{\phi}_i(v)
=
(u_i-1)\sum_{q=1}^{m_q}\omega_q\,
\mathrm{sgn}(P_q)\,\mathrm{sgn}(T_{q,i})\,
\exp\!\bigl(\log|P_q|-\log|T_{q,i}|\bigr).
\end{equation}
}

The shared quantities are computed once per node and reused
for every feature.
If all $u_j$ are nonnegative, every $T_{q,j}$ is strictly positive because $\tau_q\in(0,1)$, so sign tracking is unnecessary. For signed factors, some $T_{q,j}$ may vanish, making logarithms
and division undefined. The division-free evaluator in \Cref{subsec:prefix_suffix} handles these cases by combining the products before and after each feature, with the same asymptotic complexity.

Log-space evaluation addresses intermediate numerical range; it does not eliminate rounding error or cancellation, and reconstructed products must still fit within the floating-point range. The $\epsilon$-exactness certificate therefore remains a guarantee on quadrature error in exact arithmetic.

\paragraph{Parallel evaluation.}
The shared log-magnitudes, signs, and final weighted sums can
each be computed by repeatedly combining pairs of partial results.
For a sequence of length $k$, this \emph{tree reduction}
uses $O(k)$ total work and $O(\log k)$ parallel
time~\citep{blelloch1990prefix}.
The reductions over features can run simultaneously at all nodes,
and the subsequent sums over quadrature nodes simultaneously for all
features; since $m_q\le\lceil d/2\rceil$, both take $O(\log d)$ parallel time.


\begin{proposition}
\label{prop:complexity}
Let $d$ be the number of features and $m_q$ the number of GL nodes.
\begin{enumerate}[nosep]
  \item \emph{(Sequential.)} The estimator $\{\widehat{\phi}_i\}_{i=1}^d$
  in~\cref{eq:gl_logspace_phi} requires $O(d\,m_q)$ total work.
  \item \emph{(Parallel.)} The parallel time complexity is
  $O(\log d)$ with $O\!\bigl(d\,m_q/\log d m_q\bigr)$ processors.
\end{enumerate}
\end{proposition}

In addition to the time complexity, memory requirements depend on how the product-game components are constructed and accumulated from ML models. Section~\ref{sec:multiplicative_models} examines this issue for empirical interventional value functions and describes a blockwise implementation that avoids materializing all component-background factor tables simultaneously.

\section{Multiplicative Models and Their Induced Product Games}
\label{sec:multiplicative_models}

Section~\ref{sec:product_games_glq} developed an exact, numerically stable algorithm for a single product game. We now show that models with a multiplicative representation across features induce value functions, interventional as well as baseline, that are weighted sums of product games.


\begin{definition}
\label{def:multiplicative_model}
A predictive model $f:\R^{d}\to\R$ is \emph{multiplicative} with $p$ components if, for $c_r \in R$,
\begin{equation}
\label{eq:multiplicative_model}
f(\x)
=
\sum_{r=1}^{p} c_r \prod_{j\in\cD} u^{(r)}_j(x_j).
\end{equation}
\end{definition}

Each index $r$ identifies a multiplicative component, formed as a product of per-feature terms. When $p=1$, the model is a single product, while $p>1$ gives a finite sum of such products. The next subsection applies this representation to several model classes, where $p$ is one, the number of training points, or the number of rules. This multiplicative function could be extended to the tensor products and explain tensor networks as well. Appendix~\ref{app:tensor_networks} discusses such a case and how \QuadraSHAP{} can compute exact computations in logarithmic time. 

The SVs of $f$ depend on how the value function treats features outside the coalition. For a multiplicative model, we can preserve the product structure by retaining $u^{(r)}_j(x_j)$ for each present feature $j\in\cS$ and replacing it with an \emph{absent factor} $\tilde{u}^{(r)}_j$ for each $j\notin\cS$. These replacement factors do not depend on the observed values $\x_{\cS}$. The simplest choice is the \emph{neutral factor} $\tilde{u}^{(r)}_j=1$, which omits absent features from each product. This construction has been used for product-kernel attribution~\citep{pkex_shapley, gevaert2024unifying} and yields a weighted sum of product games as defined in \Cref{def:product_game}. Alternatively, the \emph{baseline} value function fixes a reference point $\x^b$ and sets $\tilde{u}^{(r)}_j=u^{(r)}_j(x^b_j)$, giving $v_{\x}(\cS)=f(\x_{\cS},\x^b_{\cD\setminus\cS})$.

Our focus is the \emph{interventional} value function (\Cref{sec:preliminaries}), which averages these baseline evaluations over the joint distribution of the absent features. In practice, this expectation is estimated using a background dataset, typically a subset of the training data. We show below that the resulting empirical value function retains a weighted sum-of-products representation, with one table of absent factors per background point. The neutral-factor and baseline constructions each require only a single table, so the same computational formulation covers both with $n_b=1$ (see Appendix \ref{app:value_functions}).

\begin{proposition}
\label{prop:multiplicative_to_product_games}
Let $f$ be a multiplicative model as in \cref{eq:multiplicative_model}, and let
$\mathcal{B}=\{\tilde{\x}^{(1)},\ldots,\tilde{\x}^{(n_b)}\}$ be a background dataset. Define
$u^{(r)}_j=u^{(r)}_j(x_j)$ and
$\tilde{u}^{(rb)}_j=u^{(r)}_j(\tilde{x}^{(b)}_j)$.
Then the empirical interventional value function takes the form
\begin{equation}
\label{eq:sum_of_product_games}
\hat{v}_{\x}(\cS)
=
\frac{1}{n_b}\sum_{r=1}^{p}\sum_{b=1}^{n_b}
c_r
\prod_{j\in\cS}u^{(r)}_j
\prod_{j\notin\cS}\tilde{u}^{(rb)}_j.
\end{equation}
\end{proposition}

The two sums in \cref{eq:sum_of_product_games} do not complicate the attribution: by linearity, the SVs can be computed separately for each component--background pair and then summed with the same weights. Each pair contributes a product with one factor per feature, $u^{(r)}_j$ where the feature is present and $\tilde{u}^{(rb)}_j$ where it is absent, to which the integral argument of \Cref{prop:shapley_integral_identity_product_game} applies unchanged, giving a division-free formula.

\begin{proposition}
\label{prop:gl_empirical_interventional}
Let $\hat{v}_{\x}$ be as in \cref{eq:sum_of_product_games} and $\{(\tau_q,\omega_q)\}_{q=1}^{m_q}$ be GL nodes and weights. Then, 
\begin{equation}
\label{eq:gl_empirical_interventional}
\widehat{\phi}_i(\hat{v}_{\x})
:=
\frac{1}{n_b}\sum_{r=1}^{p}\sum_{b=1}^{n_b}
c_r\bigl(u_i^{(r)}-\tilde{u}_i^{(rb)}\bigr)
\sum_{q=1}^{m_q}\omega_q
\prod_{j\neq i}
\bigl((1-\tau_q)\tilde{u}_j^{(rb)}+\tau_q u_j^{(r)}\bigr).
\end{equation}
Then $\widehat{\phi}_i(\hat{v}_{\x})=\phi_i(\hat{v}_{\x})$ for every $i\in\cD$ whenever $m_q\ge\lceil d/2\rceil$, irrespective of $p$ and $n_b$.
\end{proposition}

\Cref{eq:gl_empirical_interventional} is \cref{eq:gl_quad_shapley} with $T_{q,j}=(1-\tau_q)\tilde{u}_j+\tau_q u_j$ in place of $(1-\tau_q)+\tau_q u_j$, so the shared log-space evaluator of \Cref{subsec:stable_efficient} applies to it verbatim; where absent factors vanish, as when a rule indicator is inactive at a background point, the division-free evaluator of \Cref{subsec:prefix_suffix} is used. Because whole background rows are used, dependence among the absent features is preserved exactly; only the dependence between present and absent features is severed, and that by the interventional value function itself.

\paragraph{Time and memory complexity.}
\Cref{prop:complexity} applies once per component--background pair, giving $O(p\,n_b\,d\,m_q)$ sequential work; the pairs are independent and their aggregation is one more associative reduction, so the parallel time is $O(\log d+\log(p\,n_b))$.
For the neutral-factor and baseline value functions $n_b=1$, and the work is $O(p\,d\,m_q)$. The node budget of \Cref{prop:node_budget} carries over to \cref{eq:gl_empirical_interventional} (\Cref{app:node_budget}). 
Materializing all factor tables at once would require $O(p\,n_b\,d)$ memory.
Since the outer sum in \cref{eq:sum_of_product_games} is a reduction, the pairs can be processed in blocks of size $B$ and accumulated into the running attribution vector, which reduces peak memory to $O(Bd)$.

\paragraph{Model classes with multiplicative structure.}

\begin{table}[t]
\vspace{-1em}
\centering\small
\caption{Model classes with multiplicative structure: quantity explained, per-feature factor and number of components $p$. Notation: $s_j$ are the smooth terms of a log-link GAM ($s_j(x_j)=\beta_j x_j$ for a GLM), $\pi(\x)=\Pr(Y{=}1\mid\x)$, $h_0$ the baseline hazard, $k=\prod_j k_j$ a product kernel with coefficients $\alpha_i$, and rule $\ell$ tests the features $J_\ell$ with scores $q_{\ell j}$.}
\label{tab:model_classes}
\resizebox{0.9\linewidth}{!}{%
\begin{tabular}{@{}llll@{}}
\toprule
Model & Quantity explained & Factor $u^{(r)}_j(x_j)$ & $p$ \\
\midrule
GLM / GAM, log link      & $\E[Y\mid\x]$                         & $e^{s_j(x_j)}$                          & $1$ \\
Logistic regression      & odds $\pi(\x)/(1-\pi(\x))$            & $e^{\beta_j x_j}$                       & $1$ \\
Naive Bayes              & posterior odds                        & $p(x_j\mid Y{=}1)/p(x_j\mid Y{=}0)$     & $1$ \\
Cox proportional hazards & relative hazard $h(t\mid\x)/h_0(t)$   & $e^{\beta_j x_j}$                       & $1$ \\
Product-kernel predictor & $f(\x)=\sum_i\alpha_i\,k(\x,\x^{(i)})$ & $k_j(x_j,x^{(r)}_j)$                   & $n$ \\
Rule / tree ensemble     & $f(\x)=\sum_\ell c_\ell\prod_{j\in J_\ell}q_{\ell j}(x_j)$ & $q_{\ell j}(x_j)$, and $1$ for $j\notin J_\ell$ & number of rules \\
\bottomrule
\end{tabular}
\vspace{-1em}
}
\end{table}

\Cref{tab:model_classes} identifies the output, per-feature factors, and number of components for common model families, allowing \Cref{prop:gl_empirical_interventional} to be applied directly. Constant prefactors, including intercept and prior-odds terms, are absorbed into $c_r$.

\textbf{Output scale.}
The quantity explained is the output on which the model is multiplicative: the conditional mean of a log-link GLM or GAM, representing counts, rates, or expected costs; the posterior odds of binary logistic regression or Naive Bayes; and the relative hazard of a Cox model. For logistic regression and Naive Bayes, the choice of scale is consequential. Log-odds are additive, making interventional attribution immediate, whereas exact probability-scale Shapley computation is $\#\mathrm{P}$-hard in general, even under fully factorized data distributions~\citep{tractability_shap}. The odds scale preserves the multiplicative structure and permits exact computation through our framework. By efficiency, these SVs decompose the difference between the odds at $\x$ and the average background odds, providing an odds-scale counterpart to the log-odds explanations commonly used for classification~\citep{shap,treeshap}.

\textbf{Tree structure.}
Tree ensembles fit the same framework through their root-to-leaf rules, with node counts governed by the distinct features on each path rather than the full input dimension. The path-dependent value function, tree-specific node counts, and implementation are detailed in \Cref{app:trees}.

\section{Experiments}
\label{sec:experiments}
    
We evaluate \QuadraSHAP{} along three dimensions: the node counts needed to meet a prescribed accuracy, runtime against \textsc{PKeX-Shapley} on synthetic and text-classification tasks, and scalability to high-dimensional Cox models. Comparisons with existing tree explainers, together with additional node-budget and interventional experiments, are reported in \Cref{app:experiments}.

\subsection{Certified and Observed Quadrature Budgets}
\label{sec:5:approx}

\begin{figure}[t]
 \centering
 \includegraphics[width=.9\linewidth]{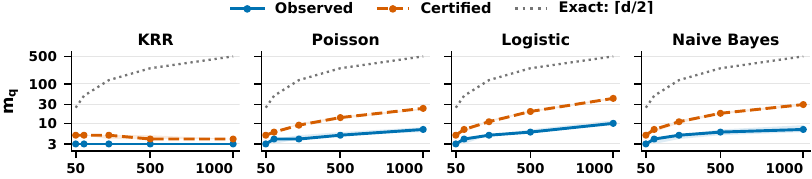}
 \caption{\textbf{Nodes needed for maximum absolute attribution error $10^{-6}$}
 versus the number of features $d$.
 Blue: \emph{observed}; orange: \emph{certified}; gray: full-dimensional
 polynomial exactness, $\lceil d/2\rceil$.
 Points are medians and bands are 10th--90th percentiles over five fits and
 20 inputs per fit.}
 \label{fig:exp8_scaling}
\end{figure}

We train RBF kernel ridge regression (KRR), Poisson regression, logistic regression, and Gaussian Naive Bayes (NB) with $d\in\{50,100,250,500,1000\}$ and $30\%$ informative features. For each model and dimension, five independent fits and 20 held-out inputs per fit give 100 explanations. We explain the KRR output, the Poisson mean, and the logistic and NB odds. For every input, we first compute the exact Shapley vector with $m_q=\lceil d/2\rceil$ nodes; the model and the input then stay fixed while $m_q$ is reduced, and the error at each $m_q$ is the largest absolute difference from the exact vector over the input's features. We compare two node counts. The \emph{observed} count is the smallest tested $m_q$ whose error is at most $\epsilon=10^{-6}$; it is known only in hindsight, because it uses the exact vector. The \emph{certified} count is the smallest $m_q$ that guarantees the given accuracy.

\Cref{fig:exp8_scaling} plots the certified and observed node counts against $d$. The certificate held for all explanations, with a worst error of $7.9\times10^{-9}$ at the certified count. At $d=1000$ the median certified counts are $4$ (KRR), $24$ (Poisson), $43$ (logistic), and $30$ (NB), against an exactness threshold of $500$; the largest count for the given accuracy over all inputs was $47$. The count that would have sufficed in hindsight is smaller, at most $14$ nodes for any input at $d=1000$. Counts grow with $d$ for Poisson, logistic, and NB but stay at $4$--$5$ for KRR at every $d$, so the feature count alone does not determine the nodes required for a given accuracy. Appendix~\ref{app:synthetic_node_budget} studies the other factors that do, including signal strength, the value function, and the variation of the factors $u_j$.

\subsection{Product-Kernel Methods Benchmark}
\label{sec:5:kernels}
We compare \QuadraSHAP{} with \textsc{PKeX-Shapley}~\citep{pkex_shapley}, the exact baseline for product-kernel models, on the same neutral-factor game $v_x(S)=\sum_r \alpha_r \prod_{j\in S}\exp[-\gamma(x_j-z_{rj})^2]$ (\Cref{app:value_functions}); the comparison with sampling estimators, which requires a background set and hence a different game, is in \Cref{app:approximate_methods}. \Cref{tab:kernel-explainer-cpu-gpu} reports \QuadraSHAP{} at the budget certified for $\epsilon=10^{-6}$ (five to six nodes) on the CPU, where \textsc{PKeX-Shapley} runs, and on the GPU. Times are median seconds per explained input; errors are the worst over the inputs.

\begin{table}[t]
\centering
\caption{Product-kernel benchmarks on synthetic and TF-IDF data. Median runtime (s/input) for the neutral-factor game, with node budgets set by the a priori quadrature tolerance $\epsilon=10^{-6}$. Small text: maximum absolute attribution error across inputs and features. \timeout{}: all inputs timed out at 300\,s. }
\label{tab:kernel-explainer-cpu-gpu}
\setlength{\tabcolsep}{3pt}
\renewcommand{\arraystretch}{1.12}
\resizebox{0.9\linewidth}{!}{%
\begin{tabular}{@{}l*{10}{r}@{}}
\toprule
& \multicolumn{5}{c}{Synthetic (varying $d$)}
& \multicolumn{5}{c}{TF-IDF text ($d=5{,}000$)} \\
\cmidrule(lr){2-6}\cmidrule(lr){7-11}
& 50 & 500 & 1k & 2k & 5k
& imdb & sms & sst2 & RT & emotion \\
\midrule
\textsc{PKeX-Shapley} (CPU, exact)
& 0.171 & 17.5 & 79.4 & \timeout & \timeout
& \timeout & \timeout & \timeout & \timeout & \timeout \\[2pt]
\textsc{QuadraSHAP} (CPU), $\epsilon{=}10^{-6}$
& 0.002 & 0.038 & 0.055 & 0.119 & 0.296
& 7.98 & 1.54 & 23.8 & 4.14 & 6.07 \\
{\scriptsize error $\le$}
& {\scriptsize $1.8\mathrm{e}{-10}$} & {\scriptsize $3.4\mathrm{e}{-10}$} & {\scriptsize $1.4\mathrm{e}{-10}$} & {\scriptsize $5.1\mathrm{e}{-11}$} & {\scriptsize $2.3\mathrm{e}{-11}$}
& {\scriptsize $4.3\mathrm{e}{-10}$} & {\scriptsize $3.9\mathrm{e}{-10}$} & {\scriptsize $1.9\mathrm{e}{-11}$} & {\scriptsize $1.3\mathrm{e}{-10}$} & {\scriptsize $9.2\mathrm{e}{-11}$} \\[2pt]
\textsc{QuadraSHAP} (GPU), $\epsilon{=}10^{-6}$
& \textbf{0.001} & \textbf{0.013} & \textbf{0.019} & \textbf{0.036} & \textbf{0.082}
& \textbf{2.12} & \textbf{0.385} & \textbf{4.93} & \textbf{1.23} & \textbf{1.35} \\
{\scriptsize error $\le$}
& {\scriptsize $1.4\mathrm{e}{-9}$} & {\scriptsize $1.7\mathrm{e}{-8}$} & {\scriptsize $3.4\mathrm{e}{-8}$} & {\scriptsize $1.9\mathrm{e}{-8}$} & {\scriptsize $8.1\mathrm{e}{-9}$}
& {\scriptsize $4.3\mathrm{e}{-10}$} & {\scriptsize $3.9\mathrm{e}{-10}$} & {\scriptsize $2.0\mathrm{e}{-11}$} & {\scriptsize $1.4\mathrm{e}{-10}$} & {\scriptsize $9.5\mathrm{e}{-11}$} \\
\bottomrule
\end{tabular}}
\end{table}


\textbf{Synthetic data.}
For each $d\in\{50,500,1000,2000,5000\}$ we generate 1{,}000 regression samples (\Cref{app:synthetic_data}), standardize them on a shuffled 80/20 split, fit RBF KRR, and explain 50 fixed held-out samples. On the same CPU, \QuadraSHAP{} takes 2\,ms at $d{=}50$ and 0.3\,s at $d{=}5000$, while \textsc{PKeX-Shapley} takes 0.17\,s, 17.5\,s, and 79.4\,s at $d=50$, $500$, and $1000$, growing quadratically with $d$ as its recursion predicts, and returns no explanation within the 300\,s limit at $d\ge2000$. On the GPU, \QuadraSHAP{} needs 1--82\,ms.

\textbf{Text classification.}
We train an RBF SVC on each of five corpora, \texttt{imdb}, \texttt{rotten tomatoes} (RT), \texttt{sst2}, \texttt{sms spam} and \texttt{emotion} (\Cref{app:datasets}), on 5{,}000 TF-IDF features, and explain 20 complete held-out texts per corpus. \QuadraSHAP{} completes every text in 1.5--24\,s on the CPU and in 0.4--4.9\,s on the GPU; \textsc{PKeX-Shapley} returns no attribution within 300\,s for any of the 100 texts, so its per-text time is above 300\,s rather than equal to it and its error is unavailable.

\subsection{Explaining High-dimensional experiments with Cox survival model}

To assess whether \QuadraSHAP{} can deliver accurate explanations at extreme dimensionality, we explain Cox models trained on GSE24080 gene expression~\citep{maqc2010microarray} and TCGA lower-grade glioma methylation~\citep{cancer2015comprehensive}, with $(n_{\mathrm{train}},d)=(339,54675)$ and $(383,396065)$, respectively. Each cohort uses 30 fixed, uniformly weighted training-background patients and five held-out patients. At a quadrature tolerance of $\epsilon=10^{-6}$, automatic node selection requires only $48$--$57$ and $41$--$53$ nodes, compared with $27{,}338$ and $198{,}033$ for full-degree quadrature. Using FP64 on an NVIDIA RTX 5090, mean evaluation times are $0.06$ and $0.36$ seconds, versus $4.81$ and $311.10$ seconds for full-degree evaluation (Table~\ref{tab:cox-cuda-compact}). The approximate timings average the median of three warm calls per patient and include automatic node selection; all timings exclude quadrature-rule construction and warmup. Maximum absolute feature-wise discrepancies from the corresponding full-degree outputs are $4.44\times10^{-10}$ and $1.75\times10^{-7}$, below the requested tolerance for every patient.

These results highlight why reducing exponential coalition enumeration to polynomial work is not sufficient by itself at this scale. Existing exact methods with $O(d^2)$ complexity substantially improve on $O(2^d)$ enumeration, yet quadratic scaling can remain costly with hundreds of thousands of features, as our own full-degree timings also illustrate. Standard Monte Carlo coalition sampling offers probabilistic error control, but conventional bounds give sufficient sample budgets that grow as $\epsilon^{-2}$ for fixed confidence and marginal-contribution range. Exploiting product structure instead allows \QuadraSHAP{} to select a sufficient budget before evaluation through a geometrically decaying, deterministic quadrature-error bound. 


\begin{table}[t]
\centering
\small
\setlength{\tabcolsep}{3pt}
\caption{Cox explanations on a GPU with 30 fixed training
backgrounds and five held-out patients per cohort. $\epsilon=0$
denotes full-degree quadrature.
$\Delta$ the maximum absolute feature-wise discrepancy across patients
and repetitions.
Times are mean $\pm$ sample SD across patient medians of three warm calls
(one call per patient for full degree).}
\resizebox{0.9\linewidth}{!}{%
\begin{tabular*}{\linewidth}{@{\extracolsep{\fill}}crrr rrr@{}}
\toprule
 & \multicolumn{3}{c}{GSE24080 (d=54,675)}
 & \multicolumn{3}{c}{TCGA LGG (d=396,065)} \\
\cmidrule(lr){2-4}\cmidrule(lr){5-7}
$\epsilon$
& Nodes & $\Delta$ & Time (s)
& Nodes & $\Delta$ & Time (s) \\
\midrule
$0$
& 27,338 & --- & $4.81\pm0.11$
& 198,033 & --- & $311.10\pm0.76$ \\
$10^{-1}$
& 44--53 & $4.51\times10^{-10}$ & $0.06\pm0.00$
& 37--49 & $1.75\times10^{-7}$ & $0.35\pm0.01$ \\
$10^{-3}$
& 46--54 & $4.48\times10^{-10}$ & $0.06\pm0.00$
& 39--51 & $1.75\times10^{-7}$ & $0.36\pm0.00$ \\
$10^{-6}$
& 48--57 & $4.44\times10^{-10}$ & $0.06\pm0.01$
& 41--53 & $1.75\times10^{-7}$ & $0.36\pm0.01$ \\
\bottomrule
\end{tabular*}
}
\label{tab:cox-cuda-compact}
\end{table}
\section{Conclusion}
\textsc{QuadraSHAP} combines efficient Shapley computation with explicit error control: a factor-based certificate selects a sufficient quadrature budget before evaluation, guaranteeing the requested attribution accuracy in exact arithmetic. This budget can be substantially smaller than that required for polynomial exactness, making low-latency explanations possible even in very high dimensions. For a Cox model with $396{,}065$ features, setting $\epsilon=10^{-6}$ reduced mean GPU evaluation time from $311$ seconds for the exact quadrature configuration to $0.36$ seconds per explanation, excluding quadrature-rule construction. These results bring large-scale attribution closer to interactive use, while retaining a principled connection between computational effort and accuracy. Floating-point rounding and background estimation remain separate error sources. Future work includes finite-precision error guarantees, feature-specific budgets, and evaluating end-to-end latency in interactive explanation workflows.



\clearpage \newpage
\bibliographystyle{iclr2027_conference}
\bibliography{references}

\clearpage \newpage

\appendix
\resumetoc
\renewcommand{\contentsname}{Appendix contents}
\setcounter{tocdepth}{2}  
\tableofcontents
\clearpage 
\section{Relation to Quadrature-Based Tree Explainers}
\label{app:related_quadrature}

Two recent methods evaluate Shapley values of decision trees by Gauss--Legendre quadrature. We summarize both, state what the three approaches share, and then list where they differ. \Cref{tab:quadrature_comparison} condenses the comparison; the tree instantiation itself is given in \Cref{app:trees} and its CPU and GPU benchmarks in \Cref{app:tree_benchmarks}.

\paragraph{TreeGrad-Shap.}
\citet{treegrad_ranker} start from Owen's multilinear extension $\bar f_{\x}$ of the tree game~\citep{owen1972multilinear} and the identity $\phi_i=\int_0^1\partial_i \bar f_{\x}(t\mathbf{1})\,dt$. Along the diagonal $t\mathbf{1}$ the gradient is a polynomial of degree at most $D-1$ for a tree of depth $D$, so a Gauss--Legendre rule with $\lceil D/2\rceil$ nodes evaluates the integral exactly, in $O(LD)$ time for a tree with $L$ leaves; the per-leaf polynomial is accumulated along the root-to-leaf path as a product of factors of the form $(1-t)+t\gamma_e$, with the factor of an ancestor divided out when a feature is split on repeatedly. The method is presented as a by-product of the paper's main contribution, feature ranking by gradient optimization, and also yields weighted Banzhaf and Beta-Shapley values. It is presented for single decision trees under the path-dependent (cover-based) value function of TreeSHAP~\citep{treeshap}; ensembles follow by linearity. Numerical stability relative to Linear TreeSHAP is a headline claim, supported by experiments; the authors note that the path products can underflow for deep trees and leave this unaddressed.

\paragraph{Quadrature-TreeSHAP.}
Concurrently with the present work, \citet{quadrature_treeshap} reformulate path-dependent TreeSHAP as integrals over $p\in[0,1]$ of ``weighted Banzhaf polynomials'': for a leaf $v$ with leaf weight $R^v_\varnothing$ (leaf value times the product of its edge probabilities) and distinct path features $M(v)$, $R^v_\varnothing\prod_{j\in\cS}(q_j-1)\prod_{j\in M(v)\setminus\cS}\bigl((1-p)+p\,q_j\bigr)$, which for $|\cS|=1$ is the integrand of \Cref{prop:shapley_integral_identity_product_game} on the leaf's product game, and for $|\cS|>1$ yields Shapley interaction indices of any order. The integrals are evaluated by Gauss--Legendre quadrature. The paper's central observation is empirical: although exactness requires $\lceil (\pathD-|\cS|+1)/2\rceil$ nodes, where $\pathD$ is the largest number of distinct features on a root-to-leaf path, a fixed budget of eight nodes reaches the numerical noise floor of the single-precision output in all their experiments, so runtime becomes practically independent of tree depth. The C++ CPU implementation has been merged into XGBoost as its TreeSHAP backend; a GPU implementation is reported separately. The method is confined to tree ensembles and the path-dependent value function.

\paragraph{What the three approaches share.}
The integral representation is the same classical identity in all three cases. For a product game the multilinear extension is $\bar v(\mathbf{z})=\prod_j\bigl((1-z_j)+z_j u_j\bigr)$, so Owen's identity yields the integral representation of \Cref{prop:shapley_integral_identity_product_game} directly; our derivation via the Beta representation of the Shapley weights arrives at the same formula. All three methods then apply Gauss--Legendre quadrature to a polynomial integrand and obtain exactness once the node count exceeds half the degree. For path-dependent trees in particular, the three methods evaluate the same per-leaf polynomial at the same nodes, at the same $O(L\,m_q)$ cost per tree; the tree instantiation of \QuadraSHAP{} should be read in that light.

\paragraph{Where they differ.}
The differences lie in the object to which the identity is applied and in the regime the resulting algorithm is built for.

\emph{Level of generality.} Both tree methods apply the identity to the tree game and are tied to the path-dependent value function. \QuadraSHAP{} applies it to the product game as an abstract object and shows in \Cref{sec:multiplicative_models} that the empirical interventional and baseline value function of any model with a sum-of-products representation is an exact weighted sum of such games. This is what carries the rule to product-kernel machines, log-link GLMs and GAMs, Cox models, odds-scale classifiers and rule ensembles, none of which is reachable from the tree formulation.

\emph{Polynomial degree.} For a tree the degree is bounded by the number of distinct features on a path, rarely more than a few tens. For the product games of \Cref{sec:multiplicative_models} it is $d_{\mathrm{eff}}$, the number of features on which the explained instance and the background differ, which for kernels and GLMs is essentially the number of features and routinely reaches thousands. Exactness then costs $O(d^2)$, so the approximation regime with provable convergence rate is a necessity rather than an optimization.

\emph{Approximation guarantee.} TreeGrad-Shap uses the exact node count only. Quadrature-TreeSHAP fixes eight nodes on empirical grounds relative to a single-precision noise floor of about $10^{-7}$; its only theoretical statement about the node count is the exactness threshold. \Cref{prop:gl_shapley_integral} gives a geometric error bound below the threshold, and \Cref{prop:node_budget} turns it into a node count computed from the factors before any evaluation. 

\emph{Dynamic range.} A product of thousands of factors can leave the range of double precision. TreeGrad-Shap acknowledges underflow for deep trees; neither tree method handles it. \QuadraSHAP{} evaluates all products in log-space with explicit sign tracking (\Cref{subsec:stable_efficient}), which is what allows the high-dimensional benchmarks of \Cref{sec:experiments}.

\emph{Parallel structure.} \QuadraSHAP{} states the computation as associative reductions and obtains $O(\log d)$ parallel time (\Cref{prop:complexity}). Quadrature-TreeSHAP provides SIMD-vectorized CPU code and a GPU implementation but does not analyze parallel time; TreeGrad-Shap does not discuss parallelism.

\emph{Interactions.} Quadrature-TreeSHAP computes Shapley interaction indices of any order for trees. The identity extends to product games in the same way, but we do not implement or evaluate interaction indices in this paper.

In summary, on path-dependent trees the first-order specializations of the three methods evaluate the same leaf-wise quadrature integrand. \Cref{app:tree_benchmarks} benchmarks \QuadraSHAP{} on CPU against the established non-quadrature tree explainers and on GPU against the XGBoost implementation of Quadrature-TreeSHAP and against GPUTreeSHAP, with the CPU implementation of TreeGrad-Shap compared separately; as discussed there, the GPU differences reflect implementation and call organization rather than a different algorithm. The contribution of \QuadraSHAP{} is the model-independent formulation, the model classes it brings within reach, and the approximation, range, and parallelism results needed once the polynomial degree is the number of features rather than the depth of a tree.

\begin{table}[h]
\centering
\footnotesize
\setlength{\tabcolsep}{4pt}
\caption{Quadrature-based Shapley methods compared. $D$ is tree depth, $\pathD$ the largest number of distinct features on a root-to-leaf path, and $d_{\mathrm{eff}}$ the number of features on which present and absent factors differ.}
\begin{tabular}{@{}>{\raggedright\arraybackslash}p{2.3cm}>{\raggedright\arraybackslash}p{3.2cm}>{\raggedright\arraybackslash}p{3.4cm}>{\raggedright\arraybackslash}p{3.9cm}@{}}
\toprule
 & \textsc{TreeGrad-Shap}~\citep{treegrad_ranker} & Quadrature-TreeSHAP~\citep{quadrature_treeshap} & \QuadraSHAP{} (this paper) \\
\midrule
Object of the identity & tree game (multilinear extension) & tree game (weighted Banzhaf polynomial) & product game, stated independently of any model \\
Value functions & path-dependent tree & path-dependent tree & empirical interventional for sum-of-products models; path-dependent tree; kernel-specific \\
Model scope & decision trees (ensembles by linearity) & tree ensembles & GLM/GAM (log link), Cox, odds-scale classifiers, product kernels, rule ensembles, trees \\
Polynomial degree & $\le$ tree depth $D$ & $\le$ distinct path features $\pathD$ & $d_{\mathrm{eff}}$, up to the number of features \\
Exact node count & $\lceil D/2\rceil$ & $\lceil (\pathD-|\cS|+1)/2\rceil$ & $\lceil d_{\mathrm{eff}}/2\rceil$ \\
Sub-threshold regime & not considered & fixed 8 nodes, empirical (single-precision floor) & geometric bound (\Cref{prop:gl_shapley_integral}); budget certified a priori from the factors (\Cref{prop:node_budget}) \\
Interaction indices & no & any order & identity stated; not implemented \\
Dynamic range & underflow noted, unhandled & not addressed & log-space with sign tracking \\
Parallelism & not discussed & SIMD CPU, GPU implementation & $O(\log d)$ parallel time via associative reductions \\
\bottomrule
\end{tabular}

\label{tab:quadrature_comparison}
\end{table}
\section{A Division-Free Variant via Prefix--Suffix Scans}  
\label{subsec:prefix_suffix}

The evaluator of \Cref{subsec:stable_efficient} forms the full product $P_q$ once and recovers every leave-one-out product by a division, which is what forces the separate treatment of vanishing factors. The leave-one-out products can instead be assembled directly from an \emph{exclusive prefix} and an \emph{exclusive suffix} product, $\prod_{j\neq i}T_{q,j}=\bigl(\prod_{j<i}T_{q,j}\bigr)\bigl(\prod_{j>i}T_{q,j}\bigr)$, which requires neither a division nor a special case. To carry this out in log-space with zeros allowed, attach to each factor the triple
\begin{equation*}
a_{q,j}:=
\begin{cases}
\bigl(\log|T_{q,j}|,\ \mathrm{sgn}(T_{q,j}),\ 0\bigr), & T_{q,j}\neq 0,\\
\bigl(0,\ 1,\ 1\bigr), & T_{q,j}=0,
\end{cases}
\qquad
(\ell_1,s_1,z_1)\oplus(\ell_2,s_2,z_2):=(\ell_1+\ell_2,\ s_1 s_2,\ z_1+z_2),
\end{equation*}
so that a triple records the log-magnitude and sign of the product of its non-zero factors together with the number of zero factors it contains. The operation $\oplus$ is associative and commutative with identity $(0,1,0)$, and a triple $(\ell,s,z)$ decodes to the product $[z=0]\,s\,e^{\ell}$. Writing $A^{<}_{q,i}:=a_{q,1}\oplus\cdots\oplus a_{q,i-1}$ and $A^{>}_{q,i}:=a_{q,i+1}\oplus\cdots\oplus a_{q,d}$ for the exclusive prefix and suffix reductions, the estimator \cref{eq:gl_quad_shapley} becomes
\begin{equation}
\label{eq:gl_prefix_suffix_phi}
\widehat{\phi}_i(v)
\;=\;
(u_i-1)\sum_{q=1}^{m_q}\omega_q\,\mathrm{dec}\bigl(A^{<}_{q,i}\oplus A^{>}_{q,i}\bigr),
\qquad
\mathrm{dec}(\ell,s,z):=[z=0]\,s\,e^{\ell}.
\end{equation}
\Cref{alg:prefix_suffix} states the procedure. All $A^{<}_{q,\cdot}$ are produced by one exclusive prefix scan of $(a_{q,j})_j$ under $\oplus$, and all $A^{>}_{q,\cdot}$ by one exclusive suffix scan, each in $O(d)$ work and $O(\log d)$ parallel time~\citep{blelloch1990prefix}; the combine and decode steps are $O(1)$ per pair $(i,q)$. \Cref{prop:complexity} therefore holds verbatim for this variant. Sequentially it makes three passes over the $d$ factors per node instead of two, a constant-factor overhead, and it performs no divisions. The same construction applies unchanged to the factors $(1-\tau_q)\tilde{u}_j+\tau_q u_j$ of \Cref{sec:multiplicative_models}.

\begin{algorithm}[t]
\caption{Division-free evaluation of \cref{eq:gl_quad_shapley} by prefix--suffix scans}
\label{alg:prefix_suffix}
\begin{algorithmic}[1]
\Require factors $u_1,\dots,u_d$; Gauss--Legendre nodes and weights $\{(\tau_q,\omega_q)\}_{q=1}^{m_q}$
\Ensure $\widehat{\phi}_1,\dots,\widehat{\phi}_d$
\State $\widehat{\phi}_i\gets 0$ for all $i$
\For{$q=1$ \textbf{to} $m_q$} \Comment{independent across $q$}
  \State $T_j\gets(1-\tau_q)+\tau_q u_j$ for all $j$ \Comment{parallel over $j$}
  \State $a_j\gets(\log|T_j|,\,\mathrm{sgn}(T_j),\,0)$ if $T_j\neq0$, else $a_j\gets(0,1,1)$, for all $j$
  \State $A^{<}\gets\textsc{ExclusivePrefixScan}\bigl((a_j)_{j=1}^d;\oplus\bigr)$ \Comment{$O(d)$ work, $O(\log d)$ parallel time}
  \State $A^{>}\gets\textsc{ExclusiveSuffixScan}\bigl((a_j)_{j=1}^d;\oplus\bigr)$
  \For{$i=1$ \textbf{to} $d$} \Comment{parallel over $i$}
    \State $(\ell,s,z)\gets A^{<}_i\oplus A^{>}_i$
    \State $\widehat{\phi}_i\gets\widehat{\phi}_i+\omega_q\,(u_i-1)\,[z=0]\,s\,e^{\ell}$
  \EndFor
\EndFor
\end{algorithmic}
\end{algorithm}

\textbf{Comparison with the shared-product evaluator.}
The two evaluators compute the same quantity whenever no factor vanishes, and their asymptotic work and parallel time coincide. They differ in three respects. The shared-product form of \Cref{subsec:stable_efficient} uses the scan primitive only as a reduction and needs one division per pair $(i,q)$; the prefix--suffix form uses two genuine scans and no division. The shared-product form must detect $\mathrm{sgn}(P_q)=0$ and recompute the affected leave-one-out products separately, whereas the prefix--suffix form treats zero factors through the counter $z$ with no change of control flow, which matters when many components have vanishing absent factors. Finally, per node and feature the shared-product form exponentiates a difference of two accumulated logarithms, while the prefix--suffix form exponentiates a single accumulated sum; the rounding behavior of the two is the same up to that last operation.
\section{Choosing the Number of Nodes for a Prescribed Accuracy}
\label{app:node_budget}

Part~(ii) of \Cref{prop:gl_shapley_integral} bounds the quadrature error below the exactness threshold by an explicit multiple of $\rho^{-2m_q}$, but it does not yet tell a user how many nodes to take. The obstacle is the factor $\max_{z\in E_\rho}|g_i(z)|$: it is the largest modulus of a polynomial of degree $d-1$ over an ellipse in the complex plane, it has no closed form, it depends on the free parameter $\rho$, and evaluating it numerically for each candidate $\rho$ would cost more than the quadrature it is meant to certify. Turning \cref{eq:gl_error_bound} into a node count therefore requires an upper bound on this maximum that is explicit, is computed from the factor table in $O(d)$ operations, and holds for every $\rho>1$ so that $\rho$ can then be chosen optimally. This appendix provides such a bound, which is \Cref{prop:node_budget} of the main text, and follows it through to a procedure for the value functions of \Cref{sec:multiplicative_models}. The maximum is bounded by a single number computed from the factor table, the total relative variation $\Lambda_i$ of \cref{eq:nb_Lambda} (\Cref{subsec:nb_ellipse}); the best $\rho$ is the root of a scalar equation (\Cref{subsec:nb_optimize}); and the bound is extended to the weighted sums of games of \Cref{sec:multiplicative_models}, where \Cref{alg:nb_budget} states the resulting procedure (\Cref{subsec:nb_sums}). The result is a node count, computed before any quadrature is run, that certifies the accuracy of all $d$ attributions at once.

\paragraph{Notation.}
Throughout, $u_j\in\R$ is the present factor of feature $j$ and $\tilde{u}_j\in\R$ its absent factor: the product game of \Cref{def:product_game} is the case $\tilde{u}_j=1$, and the component--background games of \cref{eq:sum_of_product_games} have $\tilde{u}_j=\tilde{u}^{(rb)}_j$. For each feature $j$ we write the linear factor, and for each $i$ the leave-one-out integrand, as
\begin{equation}
\label{eq:nb_integrand}
T_j(t)\;:=\;(1-t)\,\tilde{u}_j+t\,u_j,
\qquad
g_i(t)\;:=\;\prod_{j\neq i}T_j(t),
\qquad t\in[0,1].
\end{equation}
$T_j$ interpolates linearly between the absent factor at $t=0$ and the present factor at $t=1$; its value $T_j(\tau_q)$ at a node is the factor $T_{q,j}$ when $\tilde{u}_j=1$, and $g_i$ is the integrand $g_i$ of \Cref{prop:gl_shapley_integral}-(ii), written for general absent factors as in \cref{eq:gl_empirical_interventional}. By \Cref{prop:shapley_integral_identity_product_game} and the proof of \Cref{prop:gl_empirical_interventional}, the exact and the quadrature Shapley values of one game are
\begin{equation}
\label{eq:nb_phi}
\phi_i=(u_i-\tilde{u}_i)\int_0^1 g_i(t)\,dt,
\qquad
\widehat{\phi}_i=(u_i-\tilde{u}_i)\sum_{q=1}^{m}\omega_q\,g_i(\tau_q),
\end{equation}
so the error in $\phi_i$ is $|u_i-\tilde{u}_i|$ times the quadrature error of $g_i$. We work with general absent factors because that is the setting of \Cref{sec:multiplicative_models}; every statement below specializes to the product game of \Cref{prop:gl_shapley_integral} by setting $\tilde{u}_j=1$ for all $j$, in which case $|u_i-\tilde{u}_i|=|u_i-1|$ and \cref{eq:nb_gl_error} below is exactly \cref{eq:gl_error_bound}. We write $m$ for the number of nodes $m_q$.

\begin{definition}[{Bernstein ellipse for $[0,1]$}]
\label{def:nb_bernstein}
For $\rho>1$, let $E_\rho$ be the image of the circle $\{|z|=\rho\}$ under $z\mapsto\tfrac12+\tfrac14\bigl(z+z^{-1}\bigr)$. It is the ellipse with foci $0$ and $1$, semi-major axis $(\rho+\rho^{-1})/4$ and semi-minor axis $(\rho-\rho^{-1})/4$. It crosses the real axis at $-\delta$ and at $1+\delta$, where
\begin{equation}
\label{eq:nb_delta}
\delta(\rho)\;:=\;\frac{\rho+\rho^{-1}-2}{4}\;=\;\sinh^2\!\Bigl(\frac{\ln\rho}{2}\Bigr)
\end{equation}
is the distance from either focus to the nearest point of the ellipse, that is, how far $E_\rho$ extends beyond each endpoint of $[0,1]$. The ellipse collapses onto $[0,1]$ as $\rho\downarrow1$ and grows without bound as $\rho\to\infty$.
\end{definition}

The proof of \Cref{prop:gl_shapley_integral}(ii) in \Cref{apx:proofs} uses only that $g_i$ is a polynomial, hence analytic in the whole plane, so \cref{eq:gl_error_bound} holds verbatim for the general factors of \cref{eq:nb_integrand}: for every $m\ge1$ and every $\rho>1$,
\begin{equation}
\label{eq:nb_gl_error}
\bigl|\widehat{\phi}_i-\phi_i\bigr|
\;\le\;
|u_i-\tilde{u}_i|\;\frac{32}{15}\,\frac{M_i(\rho)}{1-\rho^{-2}}\;\rho^{-2m},
\qquad
M_i(\rho):=\max_{z\in E_\rho}|g_i(z)| .
\end{equation}
This is the classical bound for Gauss--Legendre quadrature of analytic integrands~\citep[Thm.~19.3]{trefethen2013atap}, transferred from $[-1,1]$ to $[0,1]$. Two features of it drive everything that follows. First, because $g_i$ is a polynomial, the bound holds for \emph{every} $\rho>1$. Second, $\rho$ is therefore a free parameter: a larger $\rho$ makes $\rho^{-2m}$ smaller but makes the ellipse larger and hence $M_i(\rho)$ larger. The node budget is decided by how this trade-off resolves, which requires knowing $M_i(\rho)$ as a function of $\rho$.

\subsection{Bounding the integrand on the ellipse}
\label{subsec:nb_ellipse}

The integrand is a product of linear factors, and a linear function is easy to maximize on an ellipse.

\begin{lemma}[One linear factor]
\label{lem:nb_factor}
Let $u,\tilde{u}\in\R$, let $T(t)=(1-t)\,\tilde{u}+t\,u$ be a factor as in \cref{eq:nb_integrand}, and let $M=\max(|u|,|\tilde{u}|)$. Then
\[
\max_{z\in E_\rho}|T(z)| \;=\; M+|u-\tilde{u}|\,\delta(\rho).
\]
\end{lemma}
\begin{proof}
If $u=\tilde{u}$ then $T$ is constant and both sides equal $M$. Otherwise $T(z)=(u-\tilde{u})(z-r)$ with the real root $r=\tilde{u}/(\tilde{u}-u)$, so $\max_{E_\rho}|T|=|u-\tilde{u}|\max_{E_\rho}|z-r|$, and it remains to find the point of $E_\rho$ farthest from $r$. Writing a point of the circle $|w|=\rho$ as $w=\rho e^{\mathrm{i}\theta}$ in the defining map $z=\tfrac12+\tfrac14(w+w^{-1})$ of \Cref{def:nb_bernstein} gives the parametrization
\[
z(\theta)=\tfrac12+a\cos\theta+\mathrm{i}\,b\sin\theta,
\qquad a=\frac{\rho+\rho^{-1}}{4},\quad b=\frac{\rho-\rho^{-1}}{4},
\qquad \theta\in[0,2\pi),
\]
with $a=\tfrac12+\delta(\rho)$ by \cref{eq:nb_delta}. Since $r$ is real,
\[
|z(\theta)-r|^2=\bigl(\tfrac12-r+a\cos\theta\bigr)^2+b^2\sin^2\theta
=(\tfrac12-r)^2+b^2+2a(\tfrac12-r)\cos\theta+(a^2-b^2)\cos^2\theta ,
\]
and $a^2-b^2=\tfrac14>0$, so the right side is a convex quadratic in $\cos\theta\in[-1,1]$ and attains its maximum at an endpoint, $\cos\theta=\pm1$. These are the two real vertices of the ellipse, $z(0)=\tfrac12+a=1+\delta$ and $z(\pi)=\tfrac12-a=-\delta$. Their distances to $r$ are $|1+\delta-r|$ and $|r+\delta|$: if $r\le\tfrac12$ the farther vertex is $1+\delta$, at distance $(1-r)+\delta=|1-r|+\delta$, and if $r\ge\tfrac12$ it is $-\delta$, at distance $r+\delta=|r|+\delta$. In both cases the farther vertex extends the farther endpoint of $[0,1]$ by $\delta$, and hence
\[
\max_{z\in E_\rho}|z-r|=\max\{|r|,|1-r|\}+\delta .
\]
Finally, $|u-\tilde{u}|\,|r|=|\tilde{u}|$ and $|u-\tilde{u}|\,|1-r|=|u|$ by the definition of $r$, so multiplying by $|u-\tilde{u}|$ gives $\max_{E_\rho}|T|=\max\{|\tilde{u}|,|u|\}+|u-\tilde{u}|\,\delta=M+|u-\tilde{u}|\,\delta$.
\end{proof}

The lemma says that on the ellipse a factor is at most its size on the interval plus its slope times the overshoot $\delta(\rho)$, and it holds for any real factors, including a vanishing $u$ or $\tilde{u}$. Multiplying over the $d-1$ factors of $g_i$ and using $1+x\le e^{x}$,
\begin{equation}
\label{eq:nb_M_bound}
M_i(\rho)
\;\le\;
\prod_{j\neq i}\bigl(M_j+|u_j-\tilde{u}_j|\,\delta(\rho)\bigr)
\;\le\;
\Pi_i\,\exp\bigl(\Lambda_i\,\delta(\rho)\bigr),
\end{equation}
where
\begin{equation}
\label{eq:nb_Lambda}
M_j:=\max(|u_j|,|\tilde{u}_j|),\qquad
\Pi_i:=\prod_{j\neq i}M_j,\qquad
\Lambda_i:=\sum_{j\neq i}\frac{|u_j-\tilde{u}_j|}{M_j},
\end{equation}
with the convention that a factor with $M_j=0$ makes the whole game vanish and is excluded. We call $\Lambda_i$ the \emph{total relative variation} of the factors: it adds up, over the features other than $i$, how much each factor changes between its present and its absent value, relative to the size of that factor. Each ratio in the sum lies in $[0,2]$, and in $[0,1]$ when $u_j$ and $\tilde{u}_j$ do not have opposite signs, as for every model class of \Cref{sec:multiplicative_models}; hence $\Lambda_i\le2(d-1)$ in general and $\Lambda_i\le d-1$ in those cases. For the product game of \Cref{def:product_game} ($\tilde{u}_j=1$, with $u_j\in\R$ arbitrary) it is $\Lambda_i=\sum_{j\neq i}|1-u_j|/\max(1,|u_j|)$. When in addition $0\le u_j\le1$, as for kernel factors, this is $\Lambda_i=\sum_{j\neq i}(1-u_j)$, and for an RBF factor $u_j=\exp(-(x_j-x'_j)^2/2\ell^2)$ the inequality $1-e^{-a}\le a$ gives $\Lambda_i\le\|\x-\x'\|^2/(2\ell^2)$. The only slack in \cref{eq:nb_M_bound} is that different factors attain their maxima at opposite vertices, and the inequality $1+x\le e^x$.

\paragraph{What $\Lambda_i$ controls.}
The ratio $|u_j-\tilde{u}_j|/M_j$ measures the change in factor $j$ between its present and its absent value relative to the magnitude of that factor, and \Cref{lem:nb_factor} says that this ratio is exactly the rate at which the factor can grow when the ellipse expands: on $E_\rho$ the factor is at most $M_j\bigl(1+\delta(\rho)\,|u_j-\tilde{u}_j|/M_j\bigr)$. When the present and absent factors are close on this relative scale, $\Lambda_i$ is small and $g_i$ grows only mildly as the ellipse expands, so a large $\rho$, and with it a fast decay $\rho^{-2m}$, is affordable. For a fixed $\rho$ the bound \cref{eq:nb_M_bound} grows as $\exp(\Lambda_i\,\delta(\rho))$; after $\rho$ is optimized in \Cref{subsec:nb_optimize}, the certified error $B(m,\Lambda_i)$ is nondecreasing in $\Lambda_i$; in the regime $m\ll\Lambda_i$ the number of nodes needed for a tolerance $\eta$ grows like $\sqrt{\Lambda_i\ln(1/\eta)}$ (\Cref{subsec:nb_optimize}), so halving $\Lambda_i$ saves roughly $30\%$ of the nodes. Two consequences follow. First, features whose factor changes little between present and absent, relative to its size, contribute little to $\Lambda_i$, and features whose factor does not change at all contribute nothing; when this holds for most features, $\Lambda_i$ is far below $d-1$ and the node count is governed by $\Lambda_i$ rather than by the degree $d-1$ of $g_i$. Second, the absolute scale of the bound is $|u_i-\tilde{u}_i|\,\Pi_i$; when every factor has magnitude at most one, as for distance-based kernels, likelihood factors and tree path factors, $\Pi_i\le1$ and is typically much smaller, which lowers the certified error further, so the same tolerance is met with fewer nodes. These statements concern the bound, not the error itself: the actual error can be much smaller than $B(m,\Lambda_i)$, and the resulting budget is sufficient rather than minimal (see the remarks at the end of \Cref{subsec:nb_sums}).

\subsection{Choosing the ellipse}
\label{subsec:nb_optimize}

Before \Cref{subsec:nb_ellipse}, the error bound \cref{eq:nb_gl_error} contained the unknown maximum $M_i(\rho)$, so the parameter $\rho$ could not be chosen: making $\rho$ larger was known to shrink $\rho^{-2m}$, but its effect on $M_i(\rho)$ was not quantified. \Cref{eq:nb_M_bound} supplies exactly this missing piece. Inserting it into \cref{eq:nb_gl_error} gives, for every $\rho>1$,
\begin{equation}
\label{eq:nb_chain}
\bigl|\widehat{\phi}_i-\phi_i\bigr|
\;\le\;
|u_i-\tilde{u}_i|\,\frac{32}{15}\,\frac{M_i(\rho)}{1-\rho^{-2}}\,\rho^{-2m}
\;\le\;
|u_i-\tilde{u}_i|\,\Pi_i\;\frac{32}{15}\,\frac{\exp\bigl(\Lambda_i\,\delta(\rho)\bigr)}{1-\rho^{-2}}\,\rho^{-2m}.
\end{equation}
The right side is an elementary function of $\rho$ in which the trade-off is in plain view: $\rho^{-2m}$ decreases with $\rho$ while $\exp(\Lambda_i\delta(\rho))$ increases, so for each pair $(m,\Lambda_i)$ there is a best $\rho$, and it depends on the factors only through the single number $\Lambda_i$. Since \cref{eq:nb_chain} holds for every $\rho>1$ and the prefactor $|u_i-\tilde{u}_i|\,\Pi_i$ does not involve $\rho$, we may take the infimum over $\rho$ and obtain
\begin{equation}
\label{eq:nb_B}
\bigl|\widehat{\phi}_i-\phi_i\bigr|\;\le\;|u_i-\tilde{u}_i|\,\Pi_i\;B(m,\Lambda_i),
\qquad
B(m,\Lambda):=\inf_{\rho>1}\ \frac{32}{15}\,\frac{\exp\bigl(\Lambda\,\delta(\rho)\bigr)\,\rho^{-2m}}{1-\rho^{-2}} .
\end{equation}
The rest of this subsection shows that the infimum is attained at a unique $\rho$, characterized by a scalar equation that is solved in a few iterations; states the resulting rule for selecting the number of nodes; and shows how the resulting budget scales with $\Lambda$.

Writing $\rho=e^{s}$ and using \cref{eq:nb_delta},
\begin{equation}
\label{eq:nb_logB}
\ln B(m,\Lambda)=\inf_{s>0}\ \Bigl[\ln\tfrac{32}{15}-\ln\bigl(1-e^{-2s}\bigr)+\Lambda\sinh^2(s/2)-2ms\Bigr].
\end{equation}
The bracket is strictly convex in $s$ (its second derivative is $\tfrac{\Lambda}{2}\cosh s+4e^{2s}/(e^{2s}-1)^2>0$), so it has a unique minimizer, characterized by
\begin{equation}
\label{eq:nb_stationarity}
\frac{\Lambda}{2}\,\sinh s \;=\; 2m+\frac{2}{e^{2s}-1}.
\end{equation}
The left side increases from $0$ and the right side decreases from $+\infty$, so \cref{eq:nb_stationarity} has exactly one root $s^\star(m,\Lambda)$, found by bisection or Newton's method in a handful of iterations, and substituting it into \cref{eq:nb_logB} gives $B(m,\Lambda)$. Evaluating $B$ for given $m$ and $\Lambda$ is therefore a scalar computation: one root find and one exponential. Two monotonicity properties will be used. $B$ is decreasing in $m$, because a larger $m$ shrinks $\rho^{-2m}$ for every $\rho$; and $B$ is nondecreasing in $\Lambda$, because $\exp(\Lambda\delta)$ increases with $\Lambda$ for every fixed $\rho$.

\paragraph{Choosing the number of nodes.}
We can now state how the node count is selected for one game and a tolerance $\epsilon>0$ on $|\widehat{\phi}_i-\phi_i|$. By \cref{eq:nb_B}, any $m$ with $|u_i-\tilde{u}_i|\,\Pi_i\,B(m,\Lambda_i)\le\epsilon$ is sufficient, and since $B$ is decreasing in $m$ the smallest such $m$ is well defined:
\begin{equation}
\label{eq:nb_mq_rule}
m_q(\epsilon)\;:=\;\min\Bigl\{\,m\ge1:\ |u_i-\tilde{u}_i|\,\Pi_i\,B(m,\Lambda_i)\le\epsilon\Bigr\}\wedge\Bigl\lceil\frac d2\Bigr\rceil ,
\end{equation}
where the cap $\lceil d/2\rceil$ is the exactness threshold of \Cref{prop:gl_shapley_integral}(i), beyond which no certificate is needed. Since $B$ is decreasing in $m$, the test $|u_i-\tilde{u}_i|\,\Pi_i\,B(m,\Lambda_i)\le\epsilon$ fails for all $m$ below $m_q(\epsilon)$ and passes for all $m$ at or above it, so $m_q(\epsilon)$ is located by bisection on the integer $m$:
\begin{enumerate}[nosep, leftmargin=*]
  \item Compute $\Pi_i$ and $\Lambda_i$ from the factor table by \cref{eq:nb_Lambda}, in $O(d)$ operations. Set $m_{\mathrm{lo}}\gets0$ (the test is regarded as failing there) and $m_{\mathrm{hi}}\gets\lceil d/2\rceil$ (where the rule is exact, so the test is regarded as passing).
  \item While $m_{\mathrm{hi}}-m_{\mathrm{lo}}>1$: let $m\gets\lfloor(m_{\mathrm{lo}}+m_{\mathrm{hi}})/2\rfloor$; solve \cref{eq:nb_stationarity} for $s^\star(m,\Lambda_i)$ and evaluate $B(m,\Lambda_i)$ by \cref{eq:nb_logB}; if $|u_i-\tilde{u}_i|\,\Pi_i\,B(m,\Lambda_i)\le\epsilon$ set $m_{\mathrm{hi}}\gets m$, otherwise set $m_{\mathrm{lo}}\gets m$.
  \item Return $m_q(\epsilon)=m_{\mathrm{hi}}$.
\end{enumerate}
The invariant is that the test fails at $m_{\mathrm{lo}}$ and passes at $m_{\mathrm{hi}}$, so on exit $m_{\mathrm{hi}}$ is the smallest passing $m$, which is the minimum in \cref{eq:nb_mq_rule}; if no $m<\lceil d/2\rceil$ passes, the procedure returns the exactness threshold. The search costs $\lceil\log_2\lceil d/2\rceil\rceil$ evaluations of $B$, each a scalar root find, and it is valid for every $m$: no restriction on $m$ relative to $\Lambda_i$ enters anywhere. This is the complete node-selection rule for a single game. The value functions of \Cref{sec:multiplicative_models} are weighted sums of many games that share one node set, and there the same rule is applied once with a common $\Lambda$ and a common scale in place of $\Lambda_i$ and $|u_i-\tilde{u}_i|\Pi_i$; \Cref{subsec:nb_sums} derives these and states the resulting procedure as \Cref{alg:nb_budget}, whose search over $m$ is the one above.

\paragraph{Scaling of the budget.}
The root of \cref{eq:nb_stationarity} has no closed form, but its asymptotics are simple. For $\sqrt{\Lambda}\lesssim m\ll\Lambda$ the root is $s^\star\approx4m/\Lambda$, the bracket in \cref{eq:nb_logB} is approximately $-4m^2/\Lambda+\ln(\Lambda/m)$, and the certified error decays like $(\Lambda/m)\,e^{-4m^2/\Lambda}$: reaching a tolerance $\eta$ costs $m\approx\tfrac12\sqrt{\Lambda\ln(1/\eta)}$ nodes up to a logarithmic correction, the square root of $\Lambda$ rather than the degree $d-1$. For $m\gg\Lambda$ the decay is super-geometric, $\exp\bigl(-2m\ln(8m/e\Lambda)\bigr)$, which is the regime that matters when $\Lambda$ is small. Numerically, the budget of \cref{eq:nb_mq_rule} lies within $5$--$18\%$ of $\tfrac12\sqrt{\Lambda\ln(1/\eta)}$ for $\Lambda\in[50,5000]$ and $\eta\in[10^{-10},10^{-3}]$.

\subsection{Weighted sums of games and the procedure}
\label{subsec:nb_sums}

The value functions of \Cref{sec:multiplicative_models} and \Cref{app:value_functions} are weighted sums $v_{\x}=\sum_{r,b}w_{rb}\,v^{(rb)}$ of games with present factors $u^{(r)}_j$ and absent factors $\tilde{u}^{(rb)}_j$, all evaluated with the same node set. Write $M^{(rb)}_j$, $\Pi^{(rb)}_i$ and $\Lambda^{(rb)}_i$ for the quantities of \cref{eq:nb_Lambda} of the pair $(r,b)$, and define the \emph{full variation}
\begin{equation}
\label{eq:nb_Lambda_bar}
\bar\Lambda\;:=\;\max_{r,b}\ \sum_{j\in\cD}\frac{|u^{(r)}_j-\tilde{u}^{(rb)}_j|}{M^{(rb)}_j},
\end{equation}
one sum per pair, so that $\Lambda^{(rb)}_i\le\bar\Lambda$ for every $i$, $r$ and $b$. By linearity of $\phi_i$ and $\widehat{\phi}_i$, the triangle inequality, \cref{eq:nb_B} for each pair, and the monotonicity of $B$ in $\Lambda$,
\begin{equation}
\label{eq:nb_sum_bound}
\begin{aligned}
\bigl|\widehat{\phi}_i(v_{\x})-\phi_i(v_{\x})\bigr|
&\;\le\;\sum_{r,b}|w_{rb}|\,\bigl|u^{(r)}_i-\tilde{u}^{(rb)}_i\bigr|\,\Pi^{(rb)}_i\,B\bigl(m,\Lambda^{(rb)}_i\bigr)\\
&\;\le\;B(m,\bar\Lambda)\sum_{r,b}|w_{rb}|\,\bigl|u^{(r)}_i-\tilde{u}^{(rb)}_i\bigr|\,\Pi^{(rb)}_i
\;=:\;A_i\,B(m,\bar\Lambda).
\end{aligned}
\end{equation}
The absolute values $|w_{rb}|$ are needed because components of opposite sign may cancel in the true error, but the guarantee cannot rely on it. With $A_{\max}:=\max_iA_i$,
\begin{equation}
\label{eq:nb_common}
\max_{i\in\cD}\bigl|\widehat{\phi}_i(v_{\x})-\phi_i(v_{\x})\bigr|\;\le\;A_{\max}\,B(m,\bar\Lambda).
\end{equation}
The node count is therefore selected by the same bisection as for one game, \cref{eq:nb_mq_rule}, with $|u_i-\tilde{u}_i|\,\Pi_i$ replaced by $A_{\max}$ and $\Lambda_i$ by $\bar\Lambda$; \Cref{alg:nb_budget} states it. All inputs are obtained in one pass over the factor tables, in $O(p\,n_b\,d)$ operations, the cost of building the tables themselves: each term of \cref{eq:nb_Lambda_bar} is a sum, and the leave-one-out products $\Pi^{(rb)}_i$ come from the prefix--suffix scan of \Cref{subsec:prefix_suffix}, which needs no division and handles zero factors. No quadrature is involved.

\begin{algorithm}[h]
\caption{Node budget for a prescribed accuracy}
\label{alg:nb_budget}
\begin{algorithmic}[1]
\Require factor tables $u^{(r)}_j$, $\tilde{u}^{(rb)}_j$ and weights $w_{rb}$ of the value function; tolerance $\epsilon>0$ on $\max_i|\widehat{\phi}_i-\phi_i|$
\Ensure a node count $m$ such that $\max_i|\widehat{\phi}_i-\phi_i|\le\epsilon$
\State compute $A_i$ for all $i$ by \cref{eq:nb_sum_bound} (prefix--suffix scan per pair) and $\bar\Lambda$ by \cref{eq:nb_Lambda_bar}; $A_{\max}\gets\max_iA_i$ \Comment{$O(p\,n_b\,d)$}
\If{$A_{\max}=0$} \Return $1$ \Comment{all attributions vanish}
\EndIf
\State $\eta\gets\epsilon/A_{\max}$;\quad $m_{\mathrm{lo}}\gets0$;\quad $m_{\mathrm{hi}}\gets\lceil d/2\rceil$ \Comment{test fails at $m_{\mathrm{lo}}$, passes at $m_{\mathrm{hi}}$}
\While{$m_{\mathrm{hi}}-m_{\mathrm{lo}}>1$}
  \State $m\gets\lfloor(m_{\mathrm{lo}}+m_{\mathrm{hi}})/2\rfloor$
  \State solve $\tfrac{\bar\Lambda}{2}\sinh s=2m+2/(e^{2s}-1)$ for $s>0$ \Comment{root of \cref{eq:nb_stationarity}}
  \State $B\gets\tfrac{32}{15}\,\exp\bigl(\bar\Lambda\sinh^2(s/2)-2ms\bigr)/(1-e^{-2s})$ \Comment{$B(m,\bar\Lambda)$ by \cref{eq:nb_logB}}
  \If{$B\le\eta$} $m_{\mathrm{hi}}\gets m$
  \Else\ $m_{\mathrm{lo}}\gets m$
  \EndIf
\EndWhile
\State \Return $m_{\mathrm{hi}}$ \Comment{equals $\lceil d/2\rceil$ if no smaller $m$ passes; exact by \Cref{prop:gl_shapley_integral}(i)}
\end{algorithmic}
\end{algorithm}

Three remarks on what the certificate says. First, it bounds the quadrature error of the estimator of \Cref{prop:gl_empirical_interventional} for the given game and background set; the statistical error of the background average and floating-point rounding are separate matters. Second, the bound is absolute error on the scale $A_{\max}$, the largest possible marginal contribution, not relative error of the attribution, which can be much smaller through cancellation. For the empirical interventional value function $w_{rb}=c_r/n_b$, so $A_i$ is an average over background points and does not grow with $n_b$. Third, the budget is sufficient rather than minimal: the common bound pairs the largest scale with the largest variation, which may belong to different pairs. When this is too conservative, the intermediate expression in \cref{eq:nb_sum_bound}, with one root of \cref{eq:nb_stationarity} per pair and per candidate $m$, can be used in the test of \Cref{alg:nb_budget} instead, at negligible cost; and the same argument with $\|A\|_2=(\sum_iA_i^2)^{1/2}$ in place of $A_{\max}$ bounds the $\ell_2$ error of the whole attribution vector.
\section{Value Functions for Multiplicative Models}
\label{app:value_functions}

This appendix collects the value functions to which \Cref{prop:gl_empirical_interventional} applies and makes precise the claim of \Cref{sec:multiplicative_models} that the neutral-factor and baseline value functions are simpler cases of the empirical interventional one: they fill in a single table of absent factors, whereas the interventional value function averages over $n_b$ such tables, one per background point.

\textbf{A common form.}
Let $f$ be a multiplicative model as in \cref{eq:multiplicative_model}. Every value function below can be written as
\begin{equation}
\label{eq:absent_factor_form}
v_{\x}(\cS)
=
\sum_{r=1}^{p}\sum_{b=1}^{n_b} w_{rb}
\prod_{j\in\cS}u^{(r)}_j
\prod_{j\notin\cS}\tilde{u}^{(rb)}_j,
\qquad
u^{(r)}_j=u^{(r)}_j(x_j),
\end{equation}
for weights $w_{rb}\in\R$ and absent factors $\tilde{u}^{(rb)}_j\in\R$ that do not depend on $\cS$. For each pair $(r,b)$ the coalition value is a product with one factor per feature, $u^{(r)}_j$ if the feature is present and $\tilde{u}^{(rb)}_j$ if it is absent. The proof of \cref{prop:gl_empirical_interventional} in \Cref{apx:proofs} uses only this structure: it treats one pair at a time, never divides by an absent factor, and combines the pairs by linearity of the Shapley value. It therefore gives, for any choice of $w_{rb}$ and $\tilde{u}^{(rb)}_j$,
\begin{equation}
\label{eq:absent_factor_shapley}
\phi_i(v_{\x})
=
\sum_{r=1}^{p}\sum_{b=1}^{n_b} w_{rb}\,\bigl(u^{(r)}_i-\tilde{u}^{(rb)}_i\bigr)
\int_0^1\prod_{j\neq i}\bigl((1-t)\,\tilde{u}^{(rb)}_j+t\,u^{(r)}_j\bigr)\,dt ,
\end{equation}
and the $m_q$-point Gauss--Legendre rule evaluates every integral exactly whenever $m_q\ge\lceil d/2\rceil$, since each integrand is a polynomial of degree at most $d-1$ in $t$. The value functions differ only in how the tables $\tilde{u}^{(rb)}$ and the weights $w_{rb}$ are filled in, and the cost of \cref{eq:absent_factor_shapley} is $O(p\,n_b\,d\,m_q)$, proportional to the number of tables.

\textbf{Empirical interventional value function.}
Let $\mathcal{B}=\{\tilde{\x}^{(1)},\ldots,\tilde{\x}^{(n_b)}\}$ be a background dataset. The interventional value function $v^{\mathrm{int}}_{\x}(\cS)=\E_{X_{\cD\setminus\cS}}[f(\x_{\cS},X_{\cD\setminus\cS})]$ of \Cref{sec:preliminaries} is estimated by replacing the expectation with the average over $\mathcal{B}$,
\begin{equation}
\label{eq:app_empirical_interventional}
\hat{v}_{\x}(\cS)
=
\frac{1}{n_b}\sum_{b=1}^{n_b} f\bigl(\x_{\cS},\tilde{\x}^{(b)}_{\cD\setminus\cS}\bigr)
=
\frac{1}{n_b}\sum_{b=1}^{n_b}\sum_{r=1}^{p} c_r
\prod_{j\in\cS}u^{(r)}_j(x_j)
\prod_{j\notin\cS}u^{(r)}_j\bigl(\tilde{x}^{(b)}_j\bigr),
\end{equation}
where the second equality substitutes \cref{eq:multiplicative_model} and uses that each factor depends on one coordinate only, so that the coordinates of the mixed point $(\x_{\cS},\tilde{\x}^{(b)}_{\cD\setminus\cS})$ separate. This is \cref{eq:absent_factor_form} with $n_b$ tables, $\tilde{u}^{(rb)}_j=u^{(r)}_j(\tilde{x}^{(b)}_j)$ and $w_{rb}=c_r/n_b$, which is \cref{prop:multiplicative_to_product_games}. Because whole background rows are used, the joint distribution of the absent features enters exactly; $\hat{v}_{\x}(\varnothing)$ is the mean prediction over $\mathcal{B}$, so by efficiency the attributions sum to $f(\x)-\frac1{n_b}\sum_b f(\tilde{\x}^{(b)})$.

\textbf{Baseline value function.}
Fix a single reference point $\x^{b}$ and set $v_{\x}(\cS)=f(\x_{\cS},\x^{b}_{\cD\setminus\cS})$. Repeating the computation in \cref{eq:app_empirical_interventional} with $\mathcal{B}=\{\x^{b}\}$ gives
\begin{equation}
\label{eq:app_baseline}
v_{\x}(\cS)
=
\sum_{r=1}^{p} c_r
\prod_{j\in\cS}u^{(r)}_j(x_j)
\prod_{j\notin\cS}u^{(r)}_j\bigl(x^{b}_j\bigr),
\end{equation}
which is \cref{eq:absent_factor_form} with one table, $\tilde{u}^{(r)}_j=u^{(r)}_j(x^{b}_j)$ and $w_r=c_r$. The empirical interventional value function is the average of $n_b$ baseline value functions, one per background point, and by linearity its Shapley values are the averages of the corresponding baseline Shapley values; the baseline value function is thus exactly the case $n_b=1$ of \cref{prop:gl_empirical_interventional}, at cost $O(p\,d\,m_q)$. Its attributions sum to $f(\x)-f(\x^{b})$.

\textbf{Neutral factor.}
The neutral factor sets $\tilde{u}^{(r)}_j\equiv1$ for every component and every feature, so that an absent feature is simply dropped from each product:
\begin{equation}
\label{eq:app_neutral}
v_{\x}(\cS)
=
\sum_{r=1}^{p} c_r\prod_{j\in\cS}u^{(r)}_j ,
\qquad
v_{\x}(\varnothing)=\sum_{r=1}^{p} c_r .
\end{equation}
This is a weighted sum of product games in the sense of \Cref{def:product_game}, with the convention $v(\varnothing)=1$ for each of them, and it is the value function for which \Cref{sec:product_games_glq} was developed: with $\tilde{u}_j=1$, \cref{eq:absent_factor_shapley} reduces to $c_r(u^{(r)}_i-1)\int_0^1\prod_{j\neq i}\bigl((1-t)+t\,u^{(r)}_j\bigr)\,dt$, the integral identity of \Cref{prop:shapley_integral_identity_product_game}, and \cref{eq:gl_quad_shapley} is its quadrature. For a product-kernel predictor $f(\x)=\sum_{i=1}^{n}\alpha_i\prod_{j}k_j(x_j,x^{(i)}_j)$ the neutral factor gives $v_{\x}(\cS)=\sum_i\alpha_i\prod_{j\in\cS}k_j(x_j,x^{(i)}_j)$, that is, the predictor evaluated with the kernel restricted to the present features, which is the value function used by \citet{pkex_shapley}; its attributions sum to $f(\x)-\sum_i\alpha_i$. Unlike the baseline value function, the neutral factor need not correspond to any reference point, since there need not be an $\x^{b}$ with $u^{(r)}_j(x^{b}_j)=1$ for all $r$ simultaneously. Algorithmically, however, it is again the case $n_b=1$ of \Cref{prop:gl_empirical_interventional}, with the single absent-factor table filled with ones, at cost $O(p\,d\,m_q)$.

\textbf{Mean absent factor.}
A fourth option sits between the baseline and the empirical interventional value functions. If the background distribution factorizes across features, then $\E\prod_{j\notin\cS}u^{(r)}_j(X_j)=\prod_{j\notin\cS}\E\,u^{(r)}_j(X_j)$, so the interventional value function is itself of the form \cref{eq:absent_factor_form} with a single table, $\tilde{u}^{(r)}_j=\E\,u^{(r)}_j(X_j)$, estimated by the per-feature averages $\frac1{n_b}\sum_b u^{(r)}_j(\tilde{x}^{(b)}_j)$. This costs $O(p\,d\,m_q)$ after an $O(p\,n_b\,d)$ pass over the background data, against $O(p\,n_b\,d\,m_q)$ for \cref{eq:app_empirical_interventional}, but it multiplies averages where \cref{eq:app_empirical_interventional} averages products: the two coincide only when the absent features are independent, and the empirical interventional value function is the one that preserves their dependence.

\begin{table}[h]
\centering
\footnotesize
\setlength{\tabcolsep}{5pt}
\caption{Value functions for a multiplicative model as instances of \cref{eq:absent_factor_form}. Each is handled by \cref{prop:gl_empirical_interventional} with the stated number of absent-factor tables, at $O(p\,d\,m_q)$ work per table; $\bar{u}^{(r)}_j=\frac1{n_b}\sum_b u^{(r)}_j(\tilde{x}^{(b)}_j)$ and $\bar{f}_{\mathcal{B}}=\frac1{n_b}\sum_b f(\tilde{\x}^{(b)})$.}
\label{tab:value_functions}
\begin{tabular}{@{}llcc@{}}
\toprule
Value function & Absent factor $\tilde{u}^{(rb)}_j$ & Tables & $\sum_i\phi_i$ \\
\midrule
Neutral factor~\citep{pkex_shapley} & $1$ & $1$ & $f(\x)-\sum_r c_r$ \\
Baseline $\x^{b}$ & $u^{(r)}_j(x^{b}_j)$ & $1$ & $f(\x)-f(\x^{b})$ \\
Mean absent factor & $\bar{u}^{(r)}_j$ & $1$ & $f(\x)-\sum_r c_r\prod_j\bar{u}^{(r)}_j$ \\
Empirical interventional & $u^{(r)}_j(\tilde{x}^{(b)}_j)$ & $n_b$ & $f(\x)-\bar{f}_{\mathcal{B}}$ \\
\bottomrule
\end{tabular}
\end{table}
\section{Extension to Tensor Networks}
\label{app:tensor_networks}

The product games of \cref{def:product_game} are the rank-one case of a broader family of multilinear models represented as \emph{tensor networks}.
Recent work computes exact Shapley values for this family~\citep{tn_shap,shap_tensor_networks}.
This appendix reviews the approach of \citet{tn_shap}, which we refer to as TN-SHAP, identifies the step at which it recovers polynomial coefficients by interpolation, and shows that the quadrature formulation of \Cref{sec:product_games_glq} replaces that step with a strictly cheaper and numerically stable alternative.
The reader is not assumed to be familiar with tensor networks, so we begin with the minimum needed.

\subsection{Tensor networks as multilinear models}
\label{app:tn_background}

A tensor network represents a high-order tensor $\cT\in\R^{n_1\times\cdots\times n_d}$ as a collection of smaller \emph{cores} joined by contractions, drawn as a graph whose nodes are cores and whose edges are shared indices.
Edges incident to a single core are \emph{physical} indices carrying the inputs; edges between cores are \emph{bond} indices whose dimensions, collected in the bond dimension $\chi$, control the trade-off between expressiveness and cost.
Contracting the network, which sums over all bond indices, is the analog of a forward pass, and for the standard topologies costs $O(\mathrm{poly}(\chi))$ rather than the $O(2^{d})$ of an explicit tensor.

Each feature is lifted to a short vector carrying a data coordinate and a constant coordinate,
\begin{equation}
\label{eq:tn_lift}
\tilde{\x}_j = \bigl[\varphi(x_j),\,1\bigr]^{\!\top},
\end{equation}
and the network defines the multilinear map $g(\tilde{\x}_1,\dots,\tilde{\x}_d)=\cT\times_1\tilde{\x}_1\cdots\times_d\tilde{\x}_d$.
Multilinearity is the property that matters here: $g$ expands as
\begin{equation}
\label{eq:tn_multilinear}
g(\x)=\sum_{\cT'\subseteq\cD} c_{\cT'}\prod_{j\in\cT'} x_j ,
\end{equation}
so that setting $x_j=1$ for $j\in\cS$ and $x_j=0$ otherwise returns $\sum_{\cT'\subseteq\cS}c_{\cT'}$.
Coalition values are therefore structured reads of the coefficient tensor, and the $2^{d}$ coalition queries of \cref{eq:shapley_def} become queries about the coefficients $c_{\cT'}$.

\textbf{TN-SHAP is model-agnostic and operates on a surrogate.}
An arbitrary predictor $f$ is not itself a tensor network. TN-SHAP therefore proceeds in two stages~\citep{tn_shap}: a tensor network $g$ is first fitted to $f$, either globally or on a local cohort around the instance to be explained, and Shapley values are then computed \emph{exactly for $g$}.
The attributions are exact with respect to the fitted surrogate rather than the deployed model, and the practitioner additionally selects the bond dimension, the feature map $\varphi$, and the fitting region.
This is the price of generality, and it is a different design point from the model-specific route taken in the body of this paper, where the multiplicative structure is read directly off the trained model.

\subsection{Diagonal selectors and the interpolation step}
\label{app:tn_interpolation}

Because the Shapley weights $\mu(|\cS|)$ depend only on coalition \emph{size}, it suffices to recover the size-aggregated coefficients $\sum_{|\cT'|=s}c_{\cT'}$ rather than all $2^{d}$ individual ones.
\citet{tn_shap} obtain these with \emph{diagonal selector} matrices
\begin{equation}
\label{eq:tn_selector}
S_j(t)=\operatorname{Diag}(t,1),
\qquad
S_j(t)\,\tilde{\x}_j
=
\bigl[t\,\varphi(x_j),\,1\bigr]^{\!\top},
\end{equation}
which scale the data coordinate of feature $j$ while preserving its constant coordinate.
At $t=1$ the feature is fully present; at $t=0$ it is replaced by its constant channel, that is, treated as absent.
Inserting $S_j(t)$ on every physical leg and contracting yields the \emph{diagonal probe}
\begin{equation}
\label{eq:tn_probe}
p(t)
=
g\bigl(S_1(t)\tilde{\x}_1,\dots,S_d(t)\tilde{\x}_d\bigr)
=
\sum_{\cT'\subseteq\cD}c_{\cT'}\,t^{|\cT'|}
=
\sum_{s=0}^{d} t^{s}\!\!\sum_{|\cT'|=s}\!\! c_{\cT'} ,
\end{equation}
a univariate polynomial whose coefficient of $t^{s}$ is exactly the size-$s$ aggregate.
Each evaluation of $p$ costs one contraction of the network.

The remaining step is to recover those coefficients. Since $p$ has degree at most $d$, TN-SHAP evaluates it at $d$ distinct probe points $\{t_1,\dots,t_{d}\}$ and solves the resulting Vandermonde system~\citep[Sec.~3.2]{tn_shap}
\begin{equation}
\label{eq:tn_vandermonde}
V\mathbf{a}=\mathbf{y},
\qquad
V_{k\ell}=t_k^{\,\ell-1},
\qquad
y_k=p(t_k),
\end{equation}
for the coefficient vector $\mathbf{a}$, after which the Shapley values follow by applying $\mu$ to the recovered aggregates.
Chebyshev--Gauss nodes and a QR-based solve are adopted in~\citet{tn_shap} to improve conditioning.
The overall cost is $d$ contractions, each $O(\mathrm{poly}(\chi))$, together with the linear solve; \citet{tn_shap} report $O(d\,\mathrm{poly}(\chi)+d^{2})$ for first-order attributions.

We note that this is the same computational mechanism used by Linear TreeSHAP~\citep{linear_treeshap}: a polynomial is evaluated at carefully chosen nodes and its coefficients are recovered by interpolation.
The instability we analyze in \Cref{subsec:stable_efficient} therefore applies here for the same reason.
Recovering monomial coefficients from point values is an inverse problem whose conditioning grows exponentially in the degree of the polynomial for \emph{any} configuration of real nodes~\citep{gautschi1975norm}; Chebyshev placement and a QR factorization improve the constant but not the rate.
In the present setting the degree is the number of features, so the difficulty grows with $d$ rather than with any property of the model.

\subsection{Quadrature in place of interpolation}
\label{app:tn_quadrature}

The diagonal probe already contains everything needed to apply the machinery of \Cref{sec:product_games_glq}, without ever forming \cref{eq:tn_vandermonde}.
The observation is that $S_j(t)$ implements exactly the Bernoulli$(t)$ coalition mixture: with the constant coordinate playing the role of the absent state, contracting with $\{S_j(t)\}_{j\in\cD}$ returns the expected coalition value when each feature is included independently with probability $t$.
Fixing feature $i$ to be present and to be absent while all other features are held at $t$ gives
\begin{equation}
\label{eq:tn_marginal_probe}
h_i(t)
=
g\bigl(\tilde{\x}_i^{\text{on}},\{S_j(t)\tilde{\x}_j\}_{j\neq i}\bigr)
-
g\bigl(\tilde{\x}_i^{\text{off}},\{S_j(t)\tilde{\x}_j\}_{j\neq i}\bigr)
=
\sum_{\cS\subseteq\cD\setminus\{i\}} t^{|\cS|}(1-t)^{\,d-1-|\cS|}\,
\Delta_i(\cS),
\end{equation}
where $\Delta_i(\cS)=v(\cS\cup\{i\})-v(\cS)$, at the cost of two contractions per value of $t$.
Applying the Beta-function identity $\mu(s)=\int_0^1 t^{s}(1-t)^{d-s-1}dt$ used in \cref{prop:shapley_integral_identity_product_game} gives
\begin{equation}
\label{eq:tn_integral}
\phi_i(v)=\int_{0}^{1} h_i(t)\,dt ,
\end{equation}
so the Shapley value of a multilinear tensor network is the integral of a univariate polynomial of degree at most $d-1$, exactly as in the rank-one case.
The integral representation itself is classical~\citep{owen1972multilinear,weber1988probabilistic}; what \cref{eq:tn_integral} shows is that the probe TN-SHAP already constructs is the integrand, so no coefficient recovery is required.

Applying the Gauss--Legendre rule of \cref{prop:gl_shapley_integral} to \cref{eq:tn_integral} yields three consequences.

\textbf{Half the contractions.}
Interpolation must determine a degree-$(d-1)$ polynomial and therefore needs $d$ distinct evaluations.
A Gaussian rule with $m_q$ nodes integrates every polynomial of degree at most $2m_q-1$ exactly, so
\begin{equation}
\label{eq:tn_half}
m_q=\Bigl\lceil \tfrac{d}{2} \Bigr\rceil
\end{equation}
nodes suffice for an exact result.
The number of network realizations required is thus halved, from $d$ to $\lceil d/2\rceil$, and each realization is the same contraction that interpolation would have performed.
This is a property of Gaussian quadrature rather than of the network: the nodes are chosen to be optimal for integration, whereas interpolation nodes must merely be distinct.

\textbf{No inverse problem.}
The quadrature estimate is a positively weighted forward sum with nodes and weights given in closed form, so the linear system \cref{eq:tn_vandermonde} disappears entirely.
The exponential conditioning discussed in \Cref{app:tn_interpolation} is a property of that system, not of the network contraction that feeds it, and removing the system removes the instability with it.
This is the same distinction drawn in \Cref{subsec:stable_efficient}: improving the accuracy of the values supplied to an ill-conditioned solve does not improve the accuracy of its output.

\textbf{A controlled approximation regime.}
Because \cref{eq:tn_integral} is an integral rather than a coefficient-recovery problem, budgets below the exactness threshold remain meaningful: the error decays geometrically in $m_q$ with the explicit bound of \cref{prop:gl_shapley_integral}, and the corollary inverting that bound selects $m_q$ for a prescribed tolerance.
Interpolation admits no analog, since evaluating a degree-$(d-1)$ polynomial at fewer than $d$ points determines it only up to aliasing.
For large $d$ this is the practically relevant regime, and it is available here at no additional cost.

\subsection{Scope of the extension}
\label{app:tn_scope}

Two qualifications delimit what the above does and does not establish.

First, the log-space evaluation of \Cref{subsec:stable_efficient} transfers verbatim only when the per-feature quantities are scalars, that is, at bond dimension one, where the integrand is a product of scalars and its logarithm is a sum.
At higher bond dimension the contraction is a chain of matrix products, for which the corresponding safeguard is per-core normalization with the accumulated scale tracked separately rather than a logarithm.
The associative-scan formulation of \cref{prop:complexity} does carry over, since matrix multiplication is associative, although the reduction is no longer commutative.

Second, the surrogate-fidelity question is untouched.
Everything above concerns the computation of Shapley values \emph{for} a given tensor network; how well that network represents the model it was fitted to is a separate matter, and it is the reason the body of this paper takes the model-specific route wherever the multiplicative structure is available directly.
The extension described here is therefore complementary: for models that are natively multilinear, or where a tensor-network surrogate is the only available option, quadrature supersedes interpolation at half the cost and without the ill-conditioning.

\section{Proofs}\label{apx:proofs}

\subsection*{Proof of Proposition \ref{prop:shapley_integral_identity_product_game}}
Start from \cref{eq:shapley_weighted_esp_form}. The Shapley weight is a Beta integral~\citep[\S5.12]{olver2010nist},
\[
\mu(s)=\frac{s!\,(d-s-1)!}{d!}=B(s+1,\,d-s)=\int_0^1 t^{s}(1-t)^{d-s-1}\,dt,\qquad s=0,\dots,d-1 .
\]
Substituting it into \cref{eq:shapley_weighted_esp_form} and exchanging the finite sum with the integral,
\[
\phi_i(v)
=(u_i-1)\int_0^1\ \sum_{\cS\subseteq\cD\setminus\{i\}}\ \prod_{j\in\cS}t\,u_j\prod_{j\notin\cS\cup\{i\}}(1-t)\,dt
=(u_i-1)\int_0^1\prod_{j\neq i}\bigl((1-t)+t\,u_j\bigr)\,dt ,
\]
because expanding the product over the $d-1$ factors $j\neq i$ produces exactly one term per subset $\cS$, with the factor $t\,u_j$ for $j\in\cS$ and $(1-t)$ for $j\notin\cS$. \qed

\subsection*{Proof of Proposition~\ref{prop:gl_shapley_integral}}

\paragraph{Part~(i): Exactness.}
The integrand $g_i$ of \Cref{prop:gl_shapley_integral}(ii) is a polynomial
of degree at most $d-1$ in $t$.
An $m_q$-point GL rule on $[0,1]$ integrates exactly every polynomial of degree
at most $2m_q-1$~\citep{davis1984methods}.
For $m_q\ge\lceil d/2\rceil$ we have $2m_q-1\ge 2\lceil d/2\rceil-1\ge d-1$,
so the quadrature sum reproduces the exact integral:
\[
\sum_{q=1}^{m_q}\omega_q\,g_i(\tau_q)
\;=\;
\int_0^1 g_i(t)\,dt.
\]
Multiplying both sides by $(u_i-1)$ and applying
Proposition~\ref{prop:shapley_integral_identity_product_game} gives
$\widehat{\phi}_i(v)=\phi_i(v)$.

\paragraph{Part~(ii): Approximation.}
With $g_i$ as in \Cref{prop:gl_shapley_integral}(ii), Proposition~\ref{prop:shapley_integral_identity_product_game} and \cref{eq:gl_quad_shapley} give
\begin{align}
\bigl|\phi_i(v)-\widehat{\phi}_i(v)\bigr|
&=
\left|
(u_i-1)\int_0^1 g_i(t)\,dt - 
(u_i-1)\sum_{q=1}^{m_q}\omega_q\,g_i(\tau_q)
\right| \cr
&=  
|u_i-1|
\left|
\int_0^1 g_i(t)\,dt
-
\sum_{q=1}^{m_q}\omega_q\,g_i(\tau_q)
\right|.
\end{align}

Now \(g_i\) is a polynomial in \(t\) of degree at most \(d-1\), hence it is
entire and in particular analytic on and inside every Bernstein ellipse
\(E_\rho\), for any \(\rho>1\). Standard Gauss--Legendre quadrature error
bounds for analytic functions on \([-1,1]\) \citep[Thm.~19.3]{trefethen2013atap},
transferred to \([0,1]\) via the affine change of variables \(x=2t-1\), imply
\[
\left|
\int_0^1 g_i(t)\,dt
-
\sum_{q=1}^{m_q}\omega_q\,g_i(\tau_q)
\right|
\le
\frac{32}{15}\,\frac{\max_{z\in E_\rho}|g_i(z)|}{(1-\rho^{-2})\,\rho^{2m_q}}.
\]
The cited theorem states the bound \(\tfrac{64}{15}\,M\rho^{-2n}/(\rho^2-1)\) for the \((n{+}1)\)-point rule on \([-1,1]\); with \(n=m_q-1\) and the factor \(\tfrac12\) from the change of variables this gives the constant above.
Substituting this bound into the previous display yields
\[
\bigl|\phi_i(v)-\widehat{\phi}_i(v)\bigr|
\le
\frac{32}{15}\,\frac{|u_i-1|}{1-\rho^{-2}}\,\max_{z\in E_\rho}|g_i(z)|\,\rho^{-2m_q}
=|u_i-1|\,C_\rho\,M_\rho\,\rho^{-2m_q},
\]
which is \cref{eq:gl_error_bound}. Since $\rho>1$ is arbitrary, the error decays geometrically in $m_q$. \qed

\subsection*{Proof of \Cref{prop:node_budget}}

Write $\delta(\rho):=(\rho+\rho^{-1}-2)/4$, the distance by which $E_\rho$ extends beyond each endpoint of $[0,1]$. By \Cref{lem:nb_factor} in \Cref{app:node_budget}, each linear factor of $g_i$ satisfies $\max_{z\in E_\rho}|(1-z)+zu_j|=M_j+|1-u_j|\,\delta(\rho)$, so
\[
M_\rho\;\le\;\prod_{j\neq i}M_j\Bigl(1+\frac{|1-u_j|}{M_j}\,\delta(\rho)\Bigr)\;\le\;\Pi_i\,e^{\Lambda_i\,\delta(\rho)}
\]
by $1+x\le e^{x}$ (\cref{eq:nb_M_bound}). Inserting this into \cref{eq:gl_error_bound} gives $|\widehat{\phi}_i-\phi_i|\le|u_i-1|\,\Pi_i\,C_\rho\,e^{\Lambda_i\delta(\rho)}\rho^{-2m_q}$ for every $\rho>1$, and taking the infimum over $\rho$ yields \cref{eq:node_budget}. For fixed $\rho$ the expression is decreasing in $m$ and nondecreasing in $\Lambda$, hence so is the infimum. With $\rho=e^{s}$ the exponent is strictly convex in $s$, so for $\Lambda_i>0$ the infimum is attained at the unique root of the scalar \cref{eq:nb_stationarity}, found by bisection or Newton's method. Because $B$ is decreasing in $m$, the test $\max_i|u_i-1|\Pi_iB(m,\Lambda_i)\le\epsilon$ is monotone in $m$, so bisection over $\{1,\dots,\lceil d/2\rceil\}$ returns the smallest $m_q$ that passes; at $m_q=\lceil d/2\rceil$ the rule is exact by part~(i) of \Cref{prop:gl_shapley_integral}. \qed


\subsection*{Proof of Proposition~\ref{prop:complexity}}

\paragraph{Sequential.}
Computing, for each $q\in\{1,\dots,m_q\}$, the factors $T_{q,j}=(1-\tau_q)+\tau_q u_j$ and the associated quantities $\log|T_{q,j}|$ and $\mathrm{sgn}(T_{q,j})$ costs $O(d\,m_q)$ arithmetic operations in total. Accumulating $\log|P_q|=\sum_j \log|T_{q,j}|$ and $\mathrm{sgn}(P_q)=\prod_j \mathrm{sgn}(T_{q,j})$ is a length-$d$ reduction for each of the $m_q$ nodes, contributing $O(d\,m_q)$ additional work. Evaluating \cref{eq:gl_logspace_phi} then requires, for each $i\in\{1,\dots,d\}$, a weighted sum of $m_q$ terms, each obtained from $\log|P_q|-\log|T_{q,i}|$ in constant time, for a further $O(d\,m_q)$ operations. The total is therefore $O(d\,m_q)$.

\paragraph{Parallel time.}
Given an associative operator $\oplus$ and a length-$d$ sequence, a tree reduction, the reduction case of the associative scan, evaluates $a_1\oplus\cdots\oplus a_d$ in parallel time $O(\log d)$ using $O(d)$ processors~\citep{blelloch1990prefix}. Both $\log|P_q|$ (under addition) and $\mathrm{sgn}(P_q)$ (under multiplication) are such reductions over the feature axis of length $d$, and the $m_q$ nodes are independent, so the shared-product phase completes in parallel time $O(\log d)$. For each $i$, evaluating \cref{eq:gl_logspace_phi} requires (i) subtracting $\log|T_{q,i}|$ from $\log|P_q|$ pointwise (parallel time $O(1)$), and (ii) a weighted sum over $q$, which is a length-$m_q$ tree reduction with parallel time $O(\log m_q)$. Since the $d$ Shapley values are independent given $\{(\log|P_q|,\mathrm{sgn}(P_q))\}_{q=1}^{m_q}$, they are evaluated in parallel across $i$ without increasing parallel time. The overall parallel time is therefore $O(\log d+\log m_q)=O(\log d)$, as $m_q\le\lceil d/2\rceil$. The required number of processors is the ratio of total work to parallel time~\citep{brent1974parallel}, giving $O\!\bigl(d\,m_q/(\log d+\log m_q)\bigr)$ processors. The exact regime $m_q=\lceil d/2\rceil$ is a special case.\qed






\subsection*{Proof of \Cref{prop:multiplicative_to_product_games}}

The empirical interventional value function replaces the expectation in $v^{\mathrm{int}}_{\x}(\cS)=\E_{X_{\cD\setminus\cS}}\bigl[f(\x_{\cS},X_{\cD\setminus\cS})\bigr]$ by the average over the background dataset $\mathcal{B}$,
\[
\hat{v}_{\x}(\cS)=\frac{1}{n_b}\sum_{b=1}^{n_b} f\bigl(\x_{\cS},\tilde{\x}^{(b)}_{\cD\setminus\cS}\bigr),
\]
where $(\x_{\cS},\tilde{\x}^{(b)}_{\cD\setminus\cS})$ denotes the point whose $j$-th coordinate is $x_j$ for $j\in\cS$ and $\tilde{x}^{(b)}_j$ for $j\notin\cS$. Fix $b$ and $\cS$ and substitute \cref{eq:multiplicative_model}. Each factor $u^{(r)}_j$ depends on the $j$-th coordinate only, so at the mixed point it evaluates to $u^{(r)}_j(x_j)=u^{(r)}_j$ if $j\in\cS$ and to $u^{(r)}_j(\tilde{x}^{(b)}_j)=\tilde{u}^{(rb)}_j$ if $j\notin\cS$. Hence
\[
f\bigl(\x_{\cS},\tilde{\x}^{(b)}_{\cD\setminus\cS}\bigr)
=
\sum_{r=1}^{p} c_r\prod_{j\in\cD}u^{(r)}_j\bigl((\x_{\cS},\tilde{\x}^{(b)}_{\cD\setminus\cS})_j\bigr)
=
\sum_{r=1}^{p} c_r\prod_{j\in\cS}u^{(r)}_j\prod_{j\notin\cS}\tilde{u}^{(rb)}_j .
\]
Averaging over $b$ and exchanging the two finite sums gives \cref{eq:sum_of_product_games}. For each pair $(r,b)$ the summand is a product with one factor per feature, $u^{(r)}_j$ if $j\in\cS$ and $\tilde{u}^{(rb)}_j$ if $j\notin\cS$, and the weights $c_r/n_b$ and the absent factors $\tilde{u}^{(rb)}_j$ do not depend on $\cS$; thus $\hat{v}_{\x}$ is a weighted sum of $p\,n_b$ such games. When every absent factor is non-zero, dividing the $(r,b)$-th summand by $\prod_{j\in\cD}\tilde{u}^{(rb)}_j$ exhibits it as the product game of \Cref{def:product_game} with factors $u^{(r)}_j/\tilde{u}^{(rb)}_j$; the general case, in which some $\tilde{u}^{(rb)}_j$ may vanish, is handled without division by \Cref{prop:gl_empirical_interventional}. \qed

\subsection*{Proof of \Cref{prop:gl_empirical_interventional}}

By linearity of the Shapley value it suffices to treat one component--background pair, that is the game $w(\cS)=\prod_{j\in\cS}u_j\prod_{j\notin\cS}\tilde{u}_j$ for arbitrary $u_j,\tilde{u}_j\in\R$, where the superscripts $(r)$ and $(rb)$ are omitted. Its marginal contributions are
\[
w(\cS\cup\{i\})-w(\cS)
=
(u_i-\tilde{u}_i)\prod_{j\in\cS}u_j\prod_{j\notin\cS\cup\{i\}}\tilde{u}_j .
\]
Substituting the Beta representation $\mu(s)=\int_0^1 t^{s}(1-t)^{d-s-1}\,dt$ of the Shapley weights into \cref{eq:shapley_def} and exchanging the finite sum with the integral,
\[
\phi_i(w)
=
(u_i-\tilde{u}_i)\int_0^1\sum_{\cS\subseteq\cD\setminus\{i\}}\ \prod_{j\in\cS}t\,u_j\prod_{j\in\cD\setminus\cS\setminus\{i\}}(1-t)\,\tilde{u}_j\,dt
=
(u_i-\tilde{u}_i)\int_0^1\prod_{j\neq i}\bigl((1-t)\tilde{u}_j+t\,u_j\bigr)\,dt ,
\]
since the subset sum is precisely the expansion of the product over the $d-1$ features $j\neq i$. No factor is inverted, so the identity holds also when some $\tilde{u}_j$ vanish. The integrand is a polynomial of degree at most $d-1$ in $t$, so the $m_q$-point Gauss--Legendre rule evaluates the integral exactly as soon as $2m_q-1\ge d-1$, that is $m_q\ge\lceil d/2\rceil$~\citep{davis1984methods}. Summing over the $p\,n_b$ pairs with the weights $c_r/n_b$ gives \cref{eq:gl_empirical_interventional}. \qed

\section{Tree-Based Models}
\label{app:trees}

This appendix collects the material on tree ensembles: the path-dependent value function and its leaf-wise product structure (\Cref{app:tree_value_function}), and the sequential evaluation and its relation to the other quadrature-based tree explainers (\Cref{app:tree_sequential}). The CPU and GPU benchmarks are in \Cref{app:tree_benchmarks}.

\subsection{Path-dependent value function}
\label{app:tree_value_function}

Consider a decision tree with leaf set $\cL$ and edge set $\cE$. Each leaf $\ell\in\cL$ stores a prediction $\nu_\ell$, and $\cE_\ell\subseteq\cE$ denotes the edges on the root-to-$\ell$ path. In the path-dependent (cover-based) value function of TreeSHAP~\citep{treeshap}, a coalition $\cS$ is treated as observed by resolving the splits on features in $\cS$ deterministically for $\x$, while splits on features outside $\cS$ are averaged with the branch probabilities. For an edge $e\in\cE$ let $p_e\in(0,1)$ be the fraction of training samples taking that edge and $\kappa(e)\in\cD$ the feature of the split, and let $c_e(\x)=1/p_e$ if $\x$ follows $e$ and $c_e(\x)=0$ otherwise. Grouping the edges of a path by feature defines the path factor
\[
q_{j,\ell}(\x)=\prod_{e\in\cE_\ell:\,\kappa(e)=j} c_e(\x),
\]
with the empty product equal to one, so that $q_{j,\ell}(\x)=1$ for every feature not tested on the path. The value function is then
\begin{equation}
\label{eq:tree_value_product_sum}
v_{\x}(\cS)=\sum_{\ell\in\cL}\nu_\ell\Bigl(\prod_{e\in\cE_\ell}p_e\Bigr)\prod_{j\in\cS}q_{j,\ell}(\x),
\end{equation}
a weighted sum of product games in the sense of \Cref{def:product_game}, one per leaf, with factors $u_j=q_{j,\ell}(\x)$ and weights $\nu_\ell\prod_{e\in\cE_\ell}p_e$; since the branch probabilities are positive, the normalization of \Cref{def:product_game} is always available here. Only the features tested on the path to $\ell$ enter the product, so the degree of the leaf's integrand is the number $\pathD_\ell$ of distinct features on that path, at most the path length and unchanged by repeated splits on the same feature; we write $\pathD=\max_\ell\pathD_\ell$. Ensembles follow by linearity.

Applying \Cref{prop:gl_shapley_integral} leaf by leaf, with $a_r(q)=1-\tau_r+\tau_r q$, $s_r(q)=(q-1)/a_r(q)$ and $G_\ell(r)=\nu_\ell\bigl(\prod_{e\in\cE_\ell}p_e\bigr)\prod_{j\in\cD}a_r\bigl(q_{j,\ell}(\x)\bigr)$,
\begin{equation}
\label{eq:tree_quadrature}
\widehat{\phi}_i=\sum_{r=1}^{m_q}\omega_r\sum_{\ell\in\cL}G_\ell(r)\,s_r\bigl(q_{i,\ell}(\x)\bigr),
\end{equation}
which is exact for $m_q\ge\lceil\pathD/2\rceil$ irrespective of the number of features $d$. The factor $s_r$ divides the contribution of feature $i$ out of the shared product $G_\ell(r)$, exactly as in \Cref{subsec:stable_efficient}; the division is safe because $a_r(q)\ge1-\tau_r>0$ for $q\ge0$.

\subsection{Sequential evaluation and relation to other quadrature-based tree explainers}
\label{app:tree_sequential}

Evaluating \cref{eq:tree_quadrature} leaf by leaf costs $O(m_q|\cL|\pathD)$, or $O(|\cL|\pathD^2)$ in the exact regime, the cost of the original TreeSHAP. The redundancy across leaves is removed by the telescoping reorganization of Linear TreeSHAP~\citep{linear_treeshap}, which carries over to the quadrature representation unchanged. Traversing an edge $e$ with $\kappa(e)=i$ changes the state of feature $i$ from $q^{-}_e$ to $q^{+}_e=q^{-}_e\,c_e(\x)$, and the leaf-wise sum in \cref{eq:tree_quadrature} can be reindexed over edges: with $H_e(r)=\sum_{\ell\text{ below }e}G_\ell(r)$ and $\Delta_e(r)=s_r(q^{+}_e)-s_r(q^{-}_e)$,
\begin{equation}
\label{eq:tree_telescoping}
\widehat{\phi}_i=\sum_{r=1}^{m_q}\omega_r\sum_{e\in\cE:\,\kappa(e)=i}H_e(r)\,\Delta_e(r).
\end{equation}
A single depth-first traversal maintains the current path state, aggregates the leaf weights $H_e(r)$ upward and applies the increment $\Delta_e(r)$ once per edge, so every edge costs $O(m_q)$ and the whole tree $O(m_q|\cL|)$, that is $O(|\cL|\pathD)$ in the exact regime; the leaf--feature table $q_{i,\ell}$ is never materialized. This is the complexity of Linear TreeSHAP without its interpolation step, which is what removes the instability documented below.

On a CPU this computation is the same for all three quadrature-based tree explainers. Our tree instantiation was developed independently of TreeGrad-Shap~\citep{treegrad_ranker}, which appeared while this work was under development, and Quadrature-TreeSHAP~\citep{quadrature_treeshap} is concurrent; all three evaluate the same leaf-wise polynomial at the same Gauss--Legendre nodes (\Cref{app:related_quadrature}), the telescoping simplification applies to each, and they differ only in the node budget (Quadrature-TreeSHAP fixes eight nodes, the other two use the exact count) and in implementation. The stability that quadrature brings over interpolation-based coefficient recovery is therefore shared by all three, and a CPU comparison among them would compare implementations of one algorithm; the CPU benchmark below is accordingly against the established tree explainers. On the GPU the implementations differ substantially. TreeGrad-Shap has no GPU implementation, and we compare our native CUDA implementation with the CUDA backend of Quadrature-TreeSHAP in XGBoost and with GPUTreeSHAP~\citep{mitchell2022gputreeshap}; since the first two evaluate the same quadrature representation, the differences between them measure the CUDA implementations rather than the method.

\begingroup
\raggedbottom
\renewcommand{\topfraction}{0.85}\renewcommand{\bottomfraction}{0.6}\renewcommand{\floatpagefraction}{0.75}\renewcommand{\textfraction}{0.1}   
\section{Experimental Details and Additional Results}
\label{app:experiments}

This appendix collects the experimental material behind \Cref{sec:experiments}:
the conventions shared by all benchmarks (\Cref{app:exp_protocol}), the
datasets and fitted models (\Cref{app:exp_data}), the certified-budget study
of \Cref{sec:5:approx} (\Cref{app:synthetic_node_budget}), the additional
comparison of product-kernel explainers under the interventional value
function (\Cref{app:approximate_methods}), and the
CPU and GPU benchmarks of the tree instantiation with their full tables
(\Cref{app:tree_benchmarks}).

\subsection{Common protocol}
\label{app:exp_protocol}

Within each benchmark every method explains the same value function on the  same fitted model and the same inputs, so the methods' outputs are estimates of one Shapley vector and differ only in accuracy and cost. Reported times are end-to-end wall-clock times of the public explanation call; model training, model conversion and explainer construction are excluded unless stated otherwise, and the statistic (median or mean over inputs, seeds or repetitions) is stated with each table. Accuracy is measured against a reference on the same game: the maximum absolute coordinate difference from independent or high-node-count reference Shapley values where such a reference is computed (\Cref{app:synthetic_node_budget,app:approximate_methods}),
and the worst-case violation of the efficiency axiom, which any exact implementation satisfies to rounding, where the models are too large for an external reference (\Cref{app:tree_benchmarks}). For \QuadraSHAP{}, the \emph{certified} node count is the one selected from the a priori bound of \Cref{app:node_budget} before any computation, and the \emph{observed} count is the smallest tested count whose measured error meets the same tolerance;
references enter only the latter. Arithmetic is float64 unless stated.

\subsection{Datasets and models}
\label{app:exp_data}
\subsubsection{Synthetic dataset construction}
\label{app:synthetic_data}

The synthetic regression benchmark uses datasets generated with \texttt{sklearn.datasets.make\_regression}.
For each feature dimensionality, we generate $n{=}1{,}000$ samples with the following fixed parameters:
\begin{itemize}
  \item \texttt{n\_informative} $= d/4$ (one-fourth of features are informative);
  \item \texttt{n\_targets} $= 1$ (single real-valued target);
  \item \texttt{noise} $= 0.1$ (small Gaussian noise added to the target);
  \item \texttt{random\_state} $= 42$ (fixed seed for reproducibility).
\end{itemize}
For the product-kernel benchmark of \Cref{sec:5:kernels}, each dataset is split 80/20 (shuffled, seed 42), features and target are standardized on the training part, and an RBF kernel ridge regressor with $\gamma=1/d$ and ridge parameter $0.1$ is fitted; the 50 explained inputs per $d$ are fixed held-out points.

\subsubsection{Text-classification datasets and models}
\label{app:datasets}

All five datasets used in the real-world tree ensemble benchmark (\Cref{app:tree_benchmarks}) and the product-kernel benchmark (\Cref{sec:5:kernels}) are loaded directly from the Hugging Face \texttt{datasets} library.
Table~\ref{tab:dataset_stats} summarizes their basic statistics.

\begin{table}[H]
\centering
\footnotesize\setlength{\tabcolsep}{5pt}
\caption{Statistics of the five text-classification datasets.}
\label{tab:dataset_stats}
\begin{tabular}{@{\hspace{5pt}}lrrrl@{}}
\toprule
Dataset & Train & Test & Classes & Task \\
\midrule
\texttt{emotion}          & 16{,}000 & 2{,}000 & 6  & Emotion recognition \\
\texttt{imdb}             & 25{,}000 & 25{,}000 & 2 & Sentiment analysis \\
\texttt{sms\_spam}        & 4{,}459  & 1{,}115 & 2  & Spam detection \\
\texttt{sst2}             & 67{,}349 & 872     & 2  & Sentiment analysis \\
\texttt{rotten\_tomatoes} & 8{,}530  & 1{,}066 & 2  & Sentiment analysis \\
\bottomrule
\end{tabular}
\end{table}

\paragraph{Text preprocessing.}
Documents are tokenized with TF-IDF using the following fixed configuration: lower-casing, English stop-word removal, unigram and bigram features ($\mathrm{ngram\_range}{=}(1,2)$), minimum document frequency $\mathrm{min\_df}{=}3$ ($\mathrm{min\_df}{=}2$ for \texttt{sms\_spam}, whose 5{,}574 messages otherwise yield fewer than 5{,}000 terms), maximum document frequency $\mathrm{max\_df}{=}0.95$, sublinear TF scaling ($\mathrm{sublinear\_tf}{=}\mathrm{True}$), and vocabulary capped at 5,000 terms ($\mathrm{max\_features}{=}5000$). This yields a $d{=}5{,}000$ dimensional TF-IDF feature vector for each document.

\paragraph{Tree ensemble models.}
For the tree ensemble benchmark, a \texttt{Random\-Forest\-Classifier} is trained on each dataset with \texttt{n\_estimators${}={}$100} (\texttt{emotion}, \texttt{imdb}, \texttt{sms\_spam}) or \texttt{300} (\texttt{sst2}, \texttt{rotten\_tomatoes}), \texttt{max\_depth${}={}$100}, \texttt{min\_samples\_leaf${}={}$1}, and \texttt{random\_state${}={}$42}. No hyperparameter search is performed for the tree models; the configuration is fixed across all datasets to ensure a fair stability comparison at large depths.

\paragraph{SVM model.}
For the product-kernel benchmark (\Cref{sec:5:kernels}) and the matched interventional comparison (\Cref{app:approximate_methods}), one support vector classifier with RBF kernel is trained per dataset on the full training split with fixed hyperparameters $C{=}2$ and $\gamma{=}$``scale''; no per-dataset tuning is performed, so that every explained text of a corpus is scored by the same model. For \texttt{emotion} the explained score is the one-vs-rest decision value of the predicted class. Held-out accuracies are 88.0\% (\texttt{imdb}), 76.0\% (\texttt{rotten\_tomatoes}), 78.1\% (\texttt{sst2}, validation split), 98.2\% (\texttt{sms\_spam}) and 89.8\% (\texttt{emotion}); they describe the models, not the explainers. Twenty fixed held-out texts per corpus are explained, complete and untruncated (\texttt{imdb} test reviews have 118--582 words, median 216).
\subsection{Synthetic experiments on certified quadrature budgets}
\label{app:synthetic_node_budget}

This section gives the protocol and additional tests for
\Cref{sec:5:approx}. The central question is practical: \emph{how many
quadrature nodes are enough to compute every Shapley value of an input to a
specified accuracy?} A model with $d$ features produces $d$ Shapley values,
but the number of integration nodes $m_q$ need not equal
$\lceil d/2\rceil$ to get a highly accurate estimation. We compare the node count selected by the mathematical
error bound (\emph{certified}) with the smallest tested count that meets the
same tolerance in the experiment (\emph{observed}).

The primary target is maximum absolute attribution error $10^{-6}$.
\Cref{app:budget_fitted} explains the measurement and its numerical validation;
\Cref{app:budget_controls} tests why the required count changes; and
\Cref{app:budget_stress} isolates the effect of factor strength in a game whose
Shapley values are known analytically.

\subsubsection{Measuring observed and certified node counts}
\label{app:budget_fitted}

\paragraph{What is held fixed, and what changes?}
For each model family and feature count
$d\in\{50,100,250,500,1000\}$, we fit a model and then explain held-out inputs.
We keep that fitted model and each input fixed while changing $m_q$.
A new feature count requires a new fit, but a new node count does not.
Exactly $k=0.30d$ coordinates are informative in the data-generating model (see Appendix~\ref{app:synthetic_data} for data-generation process);
the fitted models are not constrained to have only $k$ nonzero coefficients.

\paragraph{Two node counts for the same input.}
Let $\widehat\phi_i^{(m_q)}(x)$ be the Shapley approximation for feature $i$ using $m_q$ nodes, and let $\phi_i^{\mathrm{ref}}(x)$ be the exact computation. The measured error is
\begin{equation}
\label{eq:exp8_error}
 E_m(x)=\max_{1\leq i\leq d}
 \left|\widehat\phi_i^{(m_q)}(x)-\phi_i^{\mathrm{ref}}(x)\right|.
\end{equation}

For the \emph{certified} count, \Cref{alg:nb_budget} computes the bound $C_m(x)=A_{\max}(x)B(m,\bar\Lambda(x))$ from the fitted model's factor tables and selects the smallest integer count for which it is at most
$\epsilon$. If no smaller count suffices, it returns
$\lceil d/2\rceil$, where the polynomial quadrature is exact. This selection does not use $\phi^{\mathrm{ref}}$ or the observed errors.

For the \emph{observed} count, we evaluate the approximation with successively larger integer counts and compare it with the reference:
\begin{equation}
\label{eq:exp8_observed}
 m_{\mathrm{obs}}(x)=\min\{m\in\mathcal M(x):E_m(x)\leq\epsilon\},
\end{equation}
where $\mathcal M(x)$ is the tested consecutive-node range.
This tells us how many nodes were sufficient for that input in the measured experiment. Computing it requires the exact SV; it is not a reference-free stopping rule for new inputs. In use, the certified count can be selected without running this comparison.

Both counts concern the same model, input, absent-feature reference, and tolerance. In particular, neither curve is a competing explanation method. For example, the logistic panel of \Cref{fig:exp8_scaling} has median counts of 10 observed and 43 certified at $d=1000$: the measured errors meet the
target with substantially fewer nodes than the bound requires. The certificate is conservative, but both counts are small compared with 500-node exactness. The median is taken separately for each quantity; these are not necessarily
the counts of one particular input.

\paragraph{Repetitions and plotted summaries.}
Each configuration uses five independent data/training seeds ($0$--$4$). Each fit has a 500-point held-out test block; its first 20 inputs are explained without filtering. Each plotted model--dimension setting therefore has $5\times20=100$ explanations, grouped within five training repetitions.
Curves show medians; shaded bands and error bars show the 10th--90th percentiles across these inputs. These describe variation among explanations, not confidence intervals for an average. No hyperparameter selection uses the held-out test block.

\paragraph{Sweep and reference calculations.}
For each input, we test every integer from one through
$\min\{\lceil d/2\rceil,m_{\mathrm{cert}}(10^{-9})+5\}$, as well as the
exactness-threshold count. We retain all results and evaluate tolerances
$10^{-3}$, $10^{-5}$, $10^{-6}$, and $10^{-9}$.
The certificate uses the full-coordinate variation $\bar\Lambda$ in
\cref{eq:nb_Lambda_bar}; neither $\bar\Lambda$ nor $A_{\max}$ is fixed or
calibrated to produce a desired node count.


\paragraph{Results across inputs.}
The main figure is in \Cref{sec:5:approx}; \Cref{tab:exp8_budget_1000} adds the spread and the largest certified counts at $d=1000$. At $d=50$ the median certified count is five for all four families, and no input needs more than six. At $d=1000$ the median certified counts are 24 (Poisson), 43 (logistic), and 30 (NB), with maxima of 26, 47, and 34 over the 100 inputs of each family; KRR stays at a median of four and a maximum of four. In hindsight, no input at $d=1000$ needs more than 14 nodes to reach $\epsilon=10^{-6}$, so per input the certificate over-provisions by a factor between 1.3 and 5.1. KRR is the case where increasing $d$ does \emph{not} increase the node count: with $\gamma=1/d$ its total variation stays near $\Lambda\approx2$ at every $d$, whereas $\Lambda$ grows to 45--91 for the other three families.

\begin{table}[H]
 \centering
 \small
 \setlength{\tabcolsep}{7pt}
 \caption{Node counts at $d=1000$, $\epsilon=10^{-6}$.
 Each row summarizes 100 explanations from five fits; the largest observed count is the
 smallest $m_q$ that met the tolerance for every input, known only in hindsight. The
 full-dimensional exactness threshold is 500.}
 \label{tab:exp8_budget_1000}
 \begin{tabular}{@{}lrrrr@{}}
 \toprule
 Model & \shortstack{Largest\\observed} & \shortstack{Certified\\median}
 & \shortstack{Certified\\10th--90th} & \shortstack{Largest\\certified}\\
 \midrule
 KRR & 3 & 4 & 4--4 & 4\\
 Poisson & 10 & 24 & 22--24 & 26\\
 Logistic & 14 & 43 & 41--45 & 47\\
 Gaussian NB & 11 & 30 & 28--32 & 34\\
 \bottomrule
 \end{tabular}
\end{table}

\paragraph{Variation across fits.}
To check that the growth with $d$ is not an artifact of a single fit, we take, for each family and seed, the median certified count at $d=50$ and at $d=1000$ and record the increase. Averaged over the five seeds, the increase is 19.2 nodes for Poisson (95\% bootstrap interval $[19.0,19.4]$), 37.7 for logistic ($[36.7,38.9]$), and 24.8 for NB ($[24.6,25.0]$), the intervals obtained by resampling the five seeds 10{,}000 times. The growth is thus consistent across fits. 

\paragraph{Error curves and numerical precision.}
The node-budget figure fixes an error tolerance and asks how many quadrature nodes are sufficient. \Cref{fig:exp8_convergence} asks the complementary question: for a given node count, how large are the observed error and its certified bound? We show the results for $d=1000$. The observed errors (blue) fall below the target $10^{-6}$ before the certified bounds (orange), illustrating the conservativeness of the certified node-selection rule. As $m_q$ increases, the quadrature truncation error initially dominates and the observed error decreases rapidly. Eventually, the truncation error becomes smaller than the numerical error introduced by float64 products and summations. Additional nodes then provide little practical improvement, and the observed error approaches a numerical plateau. The gray region indicates the approximate numerical-resolution scale identified by float64 consistency and efficiency checks; it is a diagnostic, not a rigorous floating-point error bound. The plateau is model dependent: RBF factors remain in a numerically mild range, producing errors near $10^{-16}$, whereas the larger intermediate products and cancellation arising in Poisson means, logistic odds, and Naive Bayes odds lead to plateaus around $10^{-14}$--$10^{-12}$. The certified bound can continue decreasing below these plateaus because it controls quadrature truncation in exact arithmetic, not finite-precision evaluation.

\begin{figure}[H]
 \centering
 \includegraphics[width=\linewidth]{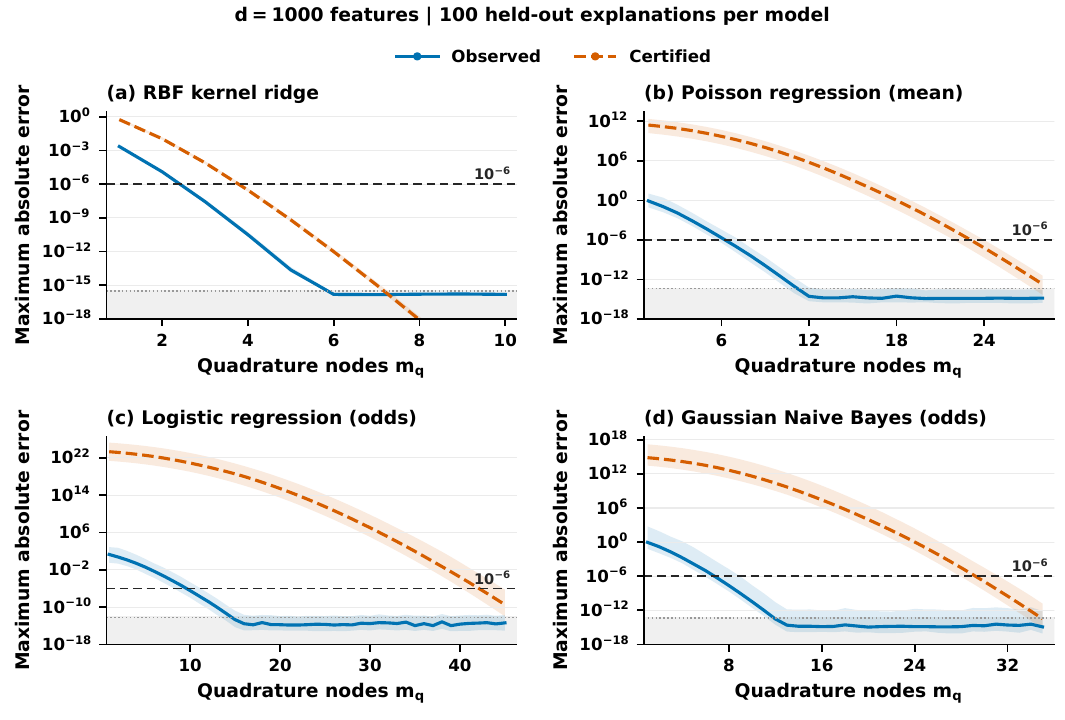}
 \caption{\textbf{Error versus node count at $d=1000$.}
The horizontal axis here is $m_q$ and the vertical axis is error. Blue: maximum absolute attribution error against the exact vector. Orange: the certified bound. Curves are medians and bands are 10th--90th percentiles over 100 explanations per family. The dashed horizontal line is the target $10^{-6}$. Panels have different vertical scales and show the consecutive-node range available for every input in that panel. Gray shading marks a median reference-discrepancy diagnostic, not a rigorous floating-point error bound.}
 \label{fig:exp8_convergence}
\end{figure}

\subsubsection{Separating feature count, signal strength, and reference choice}
\label{app:budget_controls}
\paragraph{Control 1: separating feature count from signal strength.}
In the main experiment, increasing the number of features also makes the underlying prediction problem stronger. At every dimension, $30\%$ of the features are informative, and every informative feature has a generating coefficient of magnitude $0.20$. Consequently, a model with more features also has more informative features of the same strength. Because the features are independent and standardized, the total signal in the linear predictor is measured by
\[
\|\beta\|_2^2=(0.30d)(0.20)^2=0.012d.
\]
Thus, the total signal increases from $1.2$ at $d=100$ to $12$ at $d=1000$. We call this the \emph{fixed per-feature signal} condition: each informative feature remains equally strong, so adding features increases both dimensionality and total signal.

To separate these two effects, we repeat the experiment for the three non-RBF model families while keeping the total signal constant. Specifically, we reduce the magnitude of each informative coefficient to
\[
0.20\sqrt{100/d}.
\]
There are still $0.30d$ informative features, but each feature becomes weaker as $d$ grows, giving
\[
\|\beta\|_2^2
=(0.30d)\left(0.20\sqrt{100/d}\right)^2
=1.2
\]
for every dimension. We call this the \emph{fixed total signal} condition. The two conditions are identical at $d=100$, where both use coefficient magnitude $0.20$; therefore, the repeated results at this dimension do not constitute additional independent evidence. All other data-generation, training, and explanation settings are unchanged.

The top row of \Cref{fig:exp8_controls} compares the two conditions. Solid lines represent fixed per-feature signal, while dotted lines represent fixed total signal. Blue shows the observed node count required to attain error below $10^{-6}$, and orange shows the certified node count. At $d=1000$, the observed/certified median counts decrease from $7/24$ to $4/13$ for Poisson regression, from $10/43$ to $5/23$ for logistic regression, and from $7/30$ to $5/18$ for Gaussian Naive Bayes when total signal is held fixed. Thus, the growth in node count in the main experiment is not caused by feature count alone: part of it arises because adding equally strong informative features also increases the total signal and makes the resulting product game more numerically demanding. The control does not completely isolate dimension, however, because the models are refitted at every dimension, so estimation noise, fitted coefficients, and the parameters entering the certificate may still change.

\begin{figure}[]
 \centering
 \includegraphics[width=\linewidth]{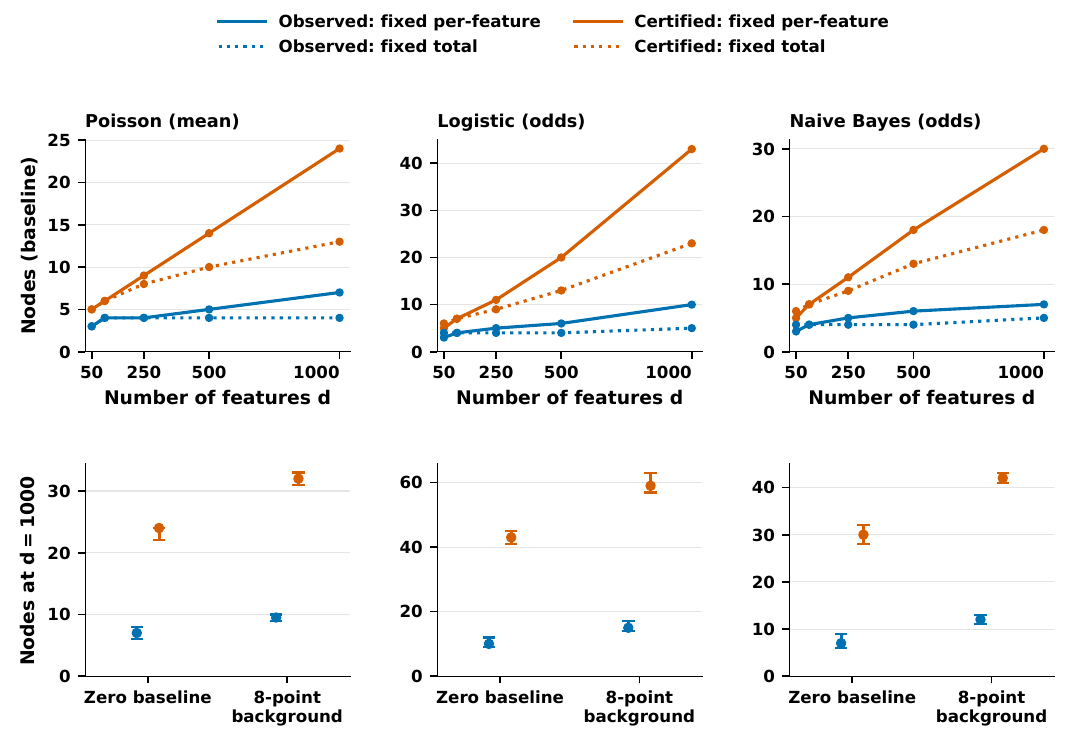}
 \caption{\textbf{Two controls for the required node count at $\epsilon=10^{-6}$.}
 Columns are Poisson mean, logistic odds, and Gaussian NB odds.
 \emph{Top:} median counts as $d$ increases, with either fixed per-feature
 signal (solid) or fixed total signal (dotted); both use a zero baseline.
 \emph{Bottom:} counts at $d=1000$ for the same main-design models and inputs,
 comparing zero replacement with an eight-point training background.
 Blue denotes observed counts and orange certified counts in both rows.
 Each setting has five fits and 20 inputs per fit; bottom-row error bars are
 10th--90th percentiles, not confidence intervals.}
 \label{fig:exp8_controls}
\end{figure}

\paragraph{Control 2: baseline versus interventional baseline.}
We next examine how the choice of value function affects the required number of quadrature nodes. Under the \emph{baseline} value function, every absent feature is replaced by zero. Under the \emph{interventional baseline}, absent features are instead replaced using eight training samples. For each of these eight samples, we evaluate the model after replacing the absent features by that sample's feature values, and then average the eight model outputs. We perform this comparison at $d=1000$ using exactly the same fitted models and test inputs. Thus, only the value function changes; the predictive model and the input being explained remain fixed. Because the two value functions define different coalition games, they can produce different Shapley values and require different quadrature budgets.

The lower row of \Cref{fig:exp8_controls} presents the results for the fixed per-feature signal condition. Moving from the baseline to the interventional baseline increases the median observed node count from $7$ to $9.5$ for Poisson regression, from $10$ to $15$ for logistic regression, and from $7$ to $12$ for Gaussian Naïve Bayes. The corresponding median certified counts increase from $24$ to $32$, from $43$ to $59$, and from $30$ to $42$, respectively. The largest certified counts under the interventional baseline are $34$, $64$, and $46$. Although the interventional value function is more demanding, these budgets remain much smaller than the exactness threshold $\lceil d/2\rceil=500$. For example, logistic regression requires a median of $59$ certified nodes, only $11.8\%$ of the exactness threshold. The value $9.5$ is possible because it is the median of an even number of integer-valued node counts.

These certificates apply to the finite interventional game defined by the eight selected training samples. They bound its quadrature error, but they do not bound the statistical error incurred by using eight samples to represent the full background population.

\subsubsection{A controlled product game with known Shapley values}
\label{app:budget_stress}

The preceding experiments use fitted models, so several quantities change simultaneously as the number of features increases, including the fitted coefficients, prediction scale, and product factors. We therefore construct a deterministic product game that isolates the effect of accumulating factor changes. Its Shapley values are available in closed form, so no model training or numerical reference calculation is required.

Let $J$ contain the $k=0.30d$ active features, while the remaining features have no effect. For a factor ratio $r\in\{1.1,1.5,3,10\}$, define
\begin{equation}
\label{eq:exp8_stress}
v(S)=r^{|S\cap J|-k},
\qquad
\phi_i=
\begin{cases}
(1-r^{-k})/k, & i\in J,\\
0, & i\notin J.
\end{cases}
\end{equation}
When no active feature is present, the coalition value is $r^{-k}$. Adding an active feature multiplies the value by $r$, and including all active features gives $v(\mathcal D)=1$. The parameter $r$ therefore controls how strongly the coalition value changes when a feature is added. Because all active features play identical roles, they divide the total change
\[
v(\mathcal D)-v(\varnothing)=1-r^{-k}
\]
equally, giving the Shapley values in \cref{eq:exp8_stress}. Inactive features have zero Shapley value.

All coalition values lie in $[0,1]$, regardless of $d$ and $r$. Consequently, any increase in the required node count cannot be attributed to an exploding output scale. For numerical stability, we implement each active feature using a present factor of $1$ and an absent factor of $1/r$, which produces exactly the game in \cref{eq:exp8_stress} while avoiding unnecessary underflow.

We evaluate the game at $d\in\{50,100,250,500,1000\}$ and target error $10^{-6}$. For each setting, we record the smallest node count that attains the target using the analytic Shapley values, together with the node count selected by the certificate.

\begin{figure}[H]
\centering
\includegraphics[width=\linewidth]{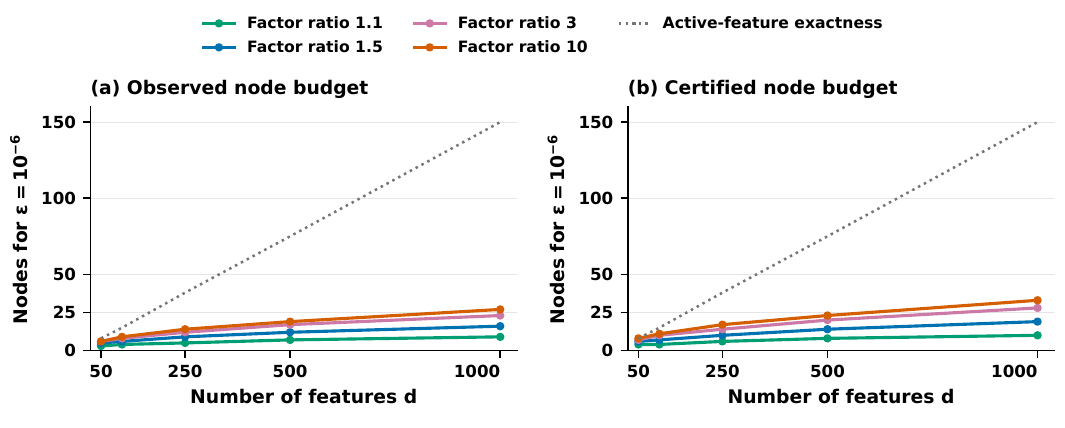}
\caption{\textbf{Node counts for the controlled product game.}
The left panel reports the observed node count required to attain error below
$10^{-6}$, using the analytic Shapley values in
\cref{eq:exp8_stress}. The right panel reports the corresponding certified
count. The horizontal axis is the number of features $d$, and colors identify
the factor ratio $r$. Only $k=0.30d$ features are active, so the relevant
exactness threshold is $\lceil k/2\rceil$, shown in gray, rather than
$\lceil d/2\rceil$. The five dimensions and four factor ratios give 20
deterministic cases; there are no training repetitions or error bars.}
\label{fig:exp8_stress}
\end{figure}

As either $d$ or $r$ increases, more active factors change in the same direction and the game requires more quadrature nodes. Formally, the amplitude term remains bounded,
\[
A_{\max}=1-\frac{1}{r},
\]
while the cumulative factor variation,
\[
\bar\Lambda=k\left(1-\frac{1}{r}\right),
\]
increases with the number of active features. Thus, the experiment increases the cumulative difficulty of the product game without increasing its output range.

For the strongest factor change, $r=10$, the observed count rises from $6$ at $d=50$ to $27$ at $d=1000$, while the certified count rises from $8$ to $33$. At $d=1000$, the observed counts for $r=1.1$, $1.5$, $3$, and $10$ are $9$, $16$, $23$, and $27$, and the corresponding certified counts are $10$, $19$, $28$, and $33$.

At $d=1000$, only $k=300$ features are active, so the exactness threshold is $\lceil k/2\rceil=150$. The largest certified budget, $33$, is therefore only $22\%$ of the exactness threshold. At the smallest setting, $d=50$ and $r=10$, the certified count equals the active-feature threshold of $8$, so the method does not provide a strict node reduction in every case. Overall, the experiment shows that the required budget can grow as more factors contribute, while still remaining much smaller than the exact quadrature budget at high dimension. This controlled result does not establish the same behavior for arbitrary signed factors, different output scalings, dimensions above $1000$, or every model and value function.

\FloatBarrier   
\subsection{Interventional explanations for product-kernel models}
\label{app:approximate_methods}

We extend the text-classification experiment in \Cref{sec:5:kernels}
to the interventional value function, comparing \QuadraSHAP{} with
\textsc{KernelSHAP} and \textsc{SamplingSHAP}~\citep{shap}.
We reuse the fitted product-kernel models and held-out texts from the main
experiment. All three methods explain the same model output using the same
background set, making their attribution errors directly comparable within
this experiment.

\paragraph{Experimental setup.}
We use the 5{,}000-feature RBF SVCs for RT, \texttt{sst2},
\texttt{sms spam}, and \texttt{emotion}. Each corpus contributes
20 held-out texts and one fixed, class-stratified background of 30 training
texts (15 per class for the binary tasks and five per class for
\texttt{emotion}), giving 80 explained texts overall. IMDB is left out because its long reviews require the sampling estimators to draw far more than 1{,}000 samples per text for a reliable Shapley estimate, which made the comparison too time-consuming, so we restrict it to the other four corpora. Each corpus uses one
fitted model; models, inputs, and backgrounds remain fixed across methods
and tolerances.

\QuadraSHAP{} runs with 
$\epsilon\in\{10^{-3},10^{-5},10^{-10},10^{-16}\}$.
The implementations of \textsc{KernelSHAP} and \textsc{SamplingSHAP}
use \textsc{shap}~\citep{shap}, the identity link, and
\texttt{nsamples=1000}. Observed error is the largest absolute attribution difference across an input's features, measured against a separate float64 \QuadraSHAP{}
calculation of the same interventional value function at a tolerance below the rounding floor. 

\begin{figure}[t]
 \centering
 \includegraphics[width=\linewidth]{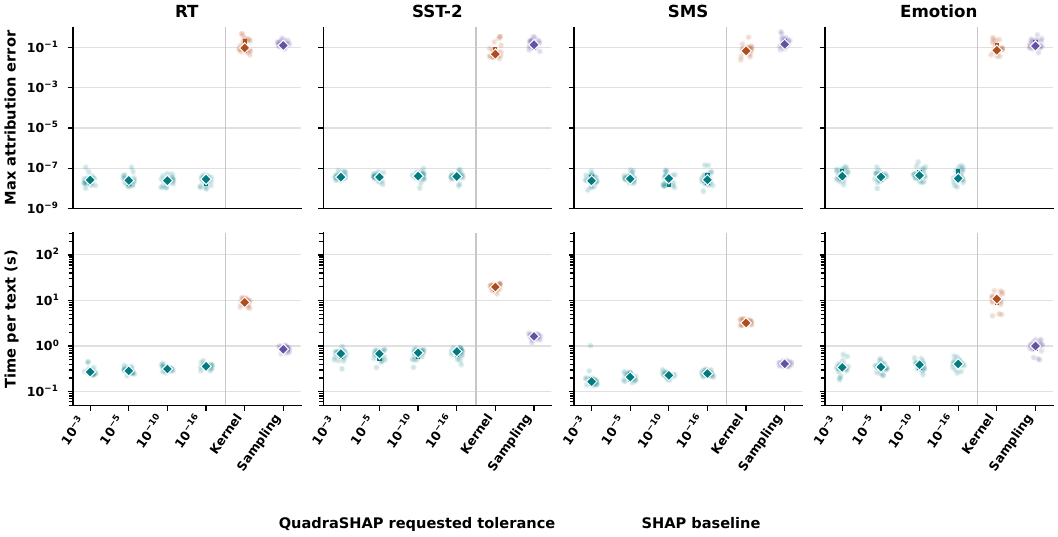}
 \caption{\textbf{Interventional attribution accuracy and runtime.}
 Top: maximum absolute attribution error per text against the float64
 calculation. Bottom: seconds per explanation. All methods use the same
 30-text background and 20 held-out inputs for each of four 5{,}000-feature
 RBF SVCs. \QuadraSHAP{} runs on a GPU in float32 at four requested tolerances;
 \textsc{KernelSHAP} and \textsc{SamplingSHAP} run on CPU with \texttt{nsamples=1000}. Faint dots are individual texts, diamonds are medians, and vertical strokes span the 25th--75th percentiles.}
 \label{fig:interventional-comparison}
\end{figure}

\paragraph{Accuracy and runtime.}
\Cref{fig:interventional-comparison} and \Cref{tab:interventional_comparison} report the results. At $\epsilon=10^{-5}$, \QuadraSHAP{} uses five nodes (six on \texttt{sst2}), takes 0.21--0.68\,s per text, and has median observed errors of $2.5$--$3.7\times10^{-8}$. \textsc{KernelSHAP} takes 3.2--19.7\,s with median errors of 0.05--0.10, and \textsc{SamplingSHAP} 0.4--1.6\,s with 0.12--0.15. These errors have to be read against the scale of the attributions: the largest attribution of a text has median magnitude 0.19--0.95 across the corpora, so the \textsc{KernelSHAP} error is 10--43\% of the largest attribution and the \textsc{SamplingSHAP} error 16--84\%, with relative $\ell_2$ errors of 0.18--0.49 and 0.29--0.94. At this budget, neither estimator resolves the attributions; \textsc{KernelSHAP} is the more accurate of the two on 61 of the 80 inputs. Therefore, \QuadraSHAP{} is not only faster, but also provides a more accurate explanation compared to the sampling methods.

\begin{table}[]
\centering
\footnotesize\setlength{\tabcolsep}{4pt}
\caption{Interventional comparison at selected tolerances, with 20 held-out
texts and one fixed 30-text background per corpus. Columns report median
node count, median time per text (seconds), and median and worst observed
maximum-coordinate error against the float64 calculation.
\QuadraSHAP{} uses the float32 GPU backend; \textsc{KernelSHAP} and \textsc{SamplingSHAP}
use CPU and \texttt{nsamples=1000}. The last column is the median of the
error divided by the largest attribution of the same text. Dashes indicate
that a quadrature node count is not applicable.}
\label{tab:interventional_comparison}
\begin{tabular}{@{}ll r r r r r@{}}
\toprule
Corpus & Method & $m_q$ & Time (s) & Median err. & Worst err. & Err.\,/\,$\max_j|\phi_j|$ \\
\midrule
RT       & \QuadraSHAP{} $10^{-3}$  & 4 & 0.27 & $2.6\times10^{-8}$ & $1.1\times10^{-7}$ & $8.1\times10^{-8}$ \\
         & \QuadraSHAP{} $10^{-5}$  & 5 & 0.29 & $2.5\times10^{-8}$ & $1.1\times10^{-7}$ & $8.5\times10^{-8}$ \\
         & \QuadraSHAP{} $10^{-10}$ & 7 & 0.31 & $2.4\times10^{-8}$ & $5.4\times10^{-8}$ & $6.7\times10^{-8}$ \\
         & \textsc{KernelSHAP}      & --& 9.07 & 0.097 & 0.442 & 0.26 \\
         & \textsc{SamplingSHAP}    & --& 0.85 & 0.129 & 0.257 & 0.54 \\
\midrule
\texttt{sst2} & \QuadraSHAP{} $10^{-3}$  & 5 & 0.68 & $3.6\times10^{-8}$ & $7.4\times10^{-8}$ & $9.2\times10^{-8}$ \\
         & \QuadraSHAP{} $10^{-5}$  & 6 & 0.68 & $3.6\times10^{-8}$ & $6.2\times10^{-8}$ & $8.7\times10^{-8}$ \\
         & \QuadraSHAP{} $10^{-10}$ & 8 & 0.71 & $4.0\times10^{-8}$ & $9.2\times10^{-8}$ & $1.0\times10^{-7}$ \\
         & \textsc{KernelSHAP}      & --& 19.72 & 0.048 & 0.311 & 0.12 \\
         & \textsc{SamplingSHAP}    & --& 1.63 & 0.140 & 0.301 & 0.38 \\
\midrule
\texttt{sms spam} & \QuadraSHAP{} $10^{-3}$  & 4 & 0.17 & $2.4\times10^{-8}$ & $6.9\times10^{-8}$ & $1.6\times10^{-7}$ \\
         & \QuadraSHAP{} $10^{-5}$  & 5 & 0.21 & $3.0\times10^{-8}$ & $7.8\times10^{-8}$ & $1.6\times10^{-7}$ \\
         & \QuadraSHAP{} $10^{-10}$ & 7 & 0.23 & $3.1\times10^{-8}$ & $7.3\times10^{-8}$ & $1.5\times10^{-7}$ \\
         & \textsc{KernelSHAP}      & --& 3.22 & 0.070 & 0.268 & 0.43 \\
         & \textsc{SamplingSHAP}    & --& 0.41 & 0.147 & 0.492 & 0.85 \\
\midrule
\texttt{emotion} & \QuadraSHAP{} $10^{-3}$  & 4 & 0.34 & $4.0\times10^{-8}$ & $1.1\times10^{-7}$ & $5.5\times10^{-8}$ \\
         & \QuadraSHAP{} $10^{-5}$  & 5 & 0.35 & $3.7\times10^{-8}$ & $1.1\times10^{-7}$ & $5.5\times10^{-8}$ \\
         & \QuadraSHAP{} $10^{-10}$ & 7 & 0.39 & $4.4\times10^{-8}$ & $2.0\times10^{-7}$ & $6.4\times10^{-8}$ \\
         & \textsc{KernelSHAP}      & --& 10.81 & 0.074 & 0.278 & 0.10 \\
         & \textsc{SamplingSHAP}    & --& 1.00 & 0.122 & 0.377 & 0.16 \\
\bottomrule
\end{tabular}
\end{table}
\FloatBarrier   
\subsection{Tree explainers}
\label{app:tree_benchmarks}

The tree instantiation of \Cref{app:trees} is benchmarked on the CPU against the established tree explainers (\Cref{app:tree_cpu_bench}) and on the GPU against Quadrature-TreeSHAP and GPUTreeSHAP, with TreeGrad-Shap compared separately (\Cref{app:tree_gpu_bench}).

\subsubsection{CPU benchmark}
\label{sec:5:trees}
\label{app:tree_cpu_bench}

We compare against \textsc{shap}~\citep{shap}, \textsc{FastTreeSHAP v1/v2}~\citep{fast_tree_shap}, \textsc{Linear TreeSHAP}~\citep{linear_treeshap, linear_treeshap_repo}, and \textsc{shapiq}~\citep{shapiq}. Numerical correctness is assessed via the worst-case violation of the Shapley efficiency axiom, which is zero by design for any exact implementation.
\begin{table}[t]
\centering
\footnotesize\setlength{\tabcolsep}{3pt}
\caption{Tree explainer benchmark: runtime (milliseconds per instance) on synthetic data across leaf counts. Stability of the efficiency axiom $\max_i|\mathbb{E}[f]+\sum_j\phi_{ij}-f(x_i)|$: \stabok{} exact (${<}10^{-12}$), \stabwarn{} degraded ($10^{-12}$ to $10^{-4}$), \stabbad{} broken (${>}10^{-4}$). Best per column \textbf{bold}; \timeout{} denotes timeout after 300s.}
\label{tab:random_treeshap_bench_simplified}
\begin{tabular}{@{}l @{\hspace{8pt}} *{5}{r} @{\hspace{10pt}} *{5}{r} @{}}
\toprule
& \multicolumn{5}{c}{$d = 10$}
& \multicolumn{5}{c}{$d = 100$} \\
\cmidrule(lr){2-6}\cmidrule(l){7-11}
\#leaves
  & $10$ & $100$ & $1\text{k}$ & $10\text{k}$ & $100\text{k}$
  & $10$ & $100$ & $1\text{k}$ & $10\text{k}$ & $100\text{k}$ \\
\midrule
\textsc{TreeSHAP}
  & 0.009\,\stabok & 0.087\,\stabok & 1.2\,\stabok & 14\,\stabwarn & 169\,\stabwarn
  & 0.010\,\stabok & 0.088\,\stabok & 1.4\,\stabok & 22.1\,\stabwarn & 254\,\stabwarn  \\
\textsc{FastTreeSHAP v1}
  & 0.010\,\stabok & 0.049\,\stabok & 0.68\,\stabok & 7.6\,\stabwarn & 89.1\,\stabwarn
  & 0.010\,\stabok & 0.055\,\stabok & 0.70\,\stabok & 10.3\,\stabwarn & 130\,\stabwarn  \\
\textsc{FastTreeSHAP v2}
  & 0.040\,\stabok & 0.11\,\stabok & 2.4\,\stabok & 7.6\,\stabwarn & 88.9\,\stabwarn
  &  0.038 \,\stabok & 0.16\,\stabok & 3.7\,\stabok & 10.6\,\stabwarn & 123\,\stabwarn  \\
\textsc{Linear TreeSHAP}
  & 0.015\,\stabok & 0.050\,\stabwarn & 0.48\,\stabwarn & 6.1\,\stabbad & 69.1\,\stabbad
  & 0.016\,\stabok & 0.050\,\stabwarn & 0.49\,\stabwarn & 5.9\,\stabbad & 66.5\,\stabbad  \\
\textsc{shapiq}
  & 3.3\,\stabok & 53.8\,\stabwarn & 573\,\stabwarn & 6281\,\stabwarn & \timeout
  & 3.4\,\stabok & 47.0\,\stabwarn & 576\,\stabwarn & 4602\,\stabwarn & 46494\,\stabbad  \\
\textbf{\QuadraSHAP}
  & \textbf{0.005}\,\stabok & \textbf{0.030}\,\stabok & \textbf{0.29}\,\stabok & \textbf{3.7}\,\stabok & \textbf{41.5}\,\stabok
  & \textbf{0.006}\,\stabok & \textbf{0.031}\,\stabok & \textbf{0.33}\,\stabok & \textbf{4.0}\,\stabok & \textbf{48.3}\,\stabok  \\
\bottomrule
\end{tabular}
\end{table}

\textbf{Synthetic data.}
We benchmark all tree-based methods on random forests trained on a controlled synthetic regression dataset (see Appendix~\ref{app:synthetic_data} for construction details), sweeping $d \in \{10, 100\}$ and ensemble size from $10$ to $10^5$ leaves. To assess algorithmic stability, we use $\max_i|\mathbb{E}[f] + \sum_j\phi_{ij} - f(x_i)|$ as a proxy, measuring the worst-case deviation from the efficiency axiom. Table~\ref{tab:random_treeshap_bench_simplified} summarizes the order-of-magnitude differences in this stability proxy; 
\textsc{QuadraSHAP} ($m_q{=}\lceil \pathD/2\rceil$) is the fastest numerically stable method in every configuration. At $10^5$ leaves it runs in 41.5\,ms ($d{=}10$) and 48.3\,ms ($d{=}100$), outperforming \textsc{FastTreeSHAP v1} by $2$--$2.7\times$ and \textsc{shap} by $3$--$5\times$, with the gap widening as the ensemble grows. Its efficiency-axiom violation remains at floating-point precision (${\sim}10^{-13}$) throughout.
\textsc{Linear TreeSHAP} matches \textsc{QuadraSHAP} in speed but is unreliable: its efficiency violation grows from $6{\times}10^{-14}$ at 10 leaves to ${\sim}10$ at $10^4$ leaves ($d{=}10$), reaching $7{\times}10^{2}$ at $10^5$ leaves ($d{=}100$). Instability sets in well before the asymptotic regime and worsens with tree depth, making its reported runtime meaningless at scale. \textsc{shapiq} is correct but orders of magnitude slower, timing out at $10^5$ leaves for $d{=}10$.

\begin{table}[t]
\centering
\footnotesize\setlength{\tabcolsep}{3pt}
\caption{Tree explainer benchmark on five text-classification datasets: runtime (millisecond per instance) with stability of the efficiency axiom shown as \stabok{} exact (${<}10^{-12}$), \stabwarn{} degraded ($10^{-12}$ to $10^{-4}$), \stabbad{} broken (${>}10^{-4}$). Best per column \textbf{bold}; \timeout{} denotes timeout after 300s.}
\label{tab:tfidf_treeshap_bench_simplified}
\begin{tabular}{@{}l r r r r r@{}}
\toprule
 & emotion & imdb & sms\_spam & sst2 & RT \\
\midrule
max depth         & 100    & 20     & 100    & 100    & 100   \\
total leaves      & 24{,}200 & 51{,}099 & 29{,}662 & 180{,}676 & 83{,}125 \\
tree count        & 100    & 100    & 100    & 300    & 300   \\
\midrule
\textsc{TreeSHAP}        & 341 \,\stabbad   & 27.4\,\stabok   & 176\,\stabbad   & 1707\,\stabbad  & 773\,\stabbad   \\
\textsc{FastTreeSHAP v1} & 326\,\stabbad   & 18.4 \,\stabok   & 166\,\stabbad   & 1621\,\stabbad  & 740\,\stabbad   \\
\textsc{FastTreeSHAP v2} & 326\,\stabbad   & \timeout   & 166\,\stabbad   & 1618\,\stabbad  & 740\,\stabbad   \\
\textsc{Linear TreeSHAP} & 43.2\,\stabbad  & 6.76\,\stabbad  & 10.8\,\stabbad  & 87.4\,\stabbad  & 
43.3\,\stabbad  \\
\textsc{shapiq}          & 1650 \,\stabbad  & 2205 \,\stabwarn & 1930 \,\stabbad  & \timeout   & \timeout  \\
\textbf{\QuadraSHAP}     & \textbf{11.3}\,\stabok & \textbf{3.75}\,\stabok & \textbf{9.68}\,\stabok & \textbf{60.5}\,\stabok & \textbf{27.3}\,\stabok \\
\bottomrule
\end{tabular}
\end{table}

\textbf{Real-world text classification.}
To assess performance on realistic, deep tree structures, we train random forest classifiers on five public text-classification datasets: \texttt{emotion}~\citep{saravia2018emotion}, \texttt{imdb}~\citep{maas2011imdb}, \texttt{sms spam}~\citep{almeida2011smsspam}, \texttt{sst2}~\citep{wang2019glue}, and \texttt{rotten tomatoes} (RT)~\citep{pang2005rottentomatoes}. 
Documents are encoded as 5,000-dimensional TF-IDF vectors; full training, and dataset details are in Appendix~\ref{app:datasets}. The resulting ensembles range from 24k to 181k leaves and reach depths up to 100, substantially deeper than synthetic forests above.

Table~\ref{tab:tfidf_treeshap_bench_simplified} reports runtime and summarizes stability differences by order of magnitude. \textsc{QuadraSHAP} is the fastest method on every dataset, achieving $2$--$4\times$ speedups over \textsc{Linear TreeSHAP} and $15$--$30\times$ over \textsc{shap} and \textsc{FastTreeSHAP}. It is also the only method that remains numerically stable across all five datasets: at depth~$100$, every competitor exhibits catastrophic efficiency violations: \textsc{shap} and \textsc{FastTreeSHAP} reach ${\sim}10^{9}$, \textsc{shapiq} reaches ${\sim}10^{30}$, and \textsc{Linear TreeSHAP} reaches $10^{186}$--$10^{271}$, while \textsc{QuadraSHAP} stays at floating-point precision (${\leq}10^{-13}$). The exception is \texttt{imdb}, whose shallow forests (max depth~$20$) fall below the breakdown threshold, and all stable methods agree there.

\subsubsection{GPU benchmark}
\label{app:tree_gpu_bench}

\textbf{Experimental setup.}
We compared the CUDA implementation of \QuadraSHAP{} with three closely related tree explainers: the GPU implementation of Quadrature-TreeSHAP in XGBoost~\citep{quadrature_treeshap}, the official \texttt{GPUTreeExplainer} of the SHAP library~\citep{mitchell2022gputreeshap}, and the official TreeGrad-Shap implementation~\citep{treegrad_ranker}. All methods computed first-order Shapley values under the path-dependent, cover-based tree value function of \Cref{app:tree_value_function}.
The primary GPU benchmark was conducted on an NVIDIA RTX A6000. We used the \QuadraSHAP{} repository at commit \texttt{3aac1099ac26a59a42acca8f12fdfe6b320cafff}, XGBoost~3.4.1 and SHAP~v0.52.0. The XGBoost binary was verified to use its Quadrature-TreeSHAP CUDA backend for \texttt{pred\_contribs=True}. Only build-compatibility changes were required to compile SHAP for the available CUDA version; the GPUTreeSHAP algorithm was not modified.
Synthetic regression data were generated with target
\[
y=\sum_{j=1}^{\min\{8,d\}}\bigl[\sin(x_j)+0.35\,x_j^{2}\bigr]+\xi,\qquad \xi\sim\mathcal{N}(0,1).
\]
The benchmark contained six model configurations: single trees with 16 features and 256, 4{,}096 or 16{,}384 leaves; a single tree with 100 features and 4{,}096 leaves; and forests of 32 trees with 128 leaves per tree and either 16 or 100 features. Tree depth was capped at 24 to remain within the path-size limitation of the released GPUTreeSHAP implementation, and every requested leaf count was attained. Each configuration was trained with three seeds (42, 43, 44) and evaluated at batch sizes $B\in\{1,10,64\}$, giving 18 model--batch cells. The same fitted tree topology, thresholds, leaf values, covers and input observations were supplied to all methods. Conversion to XGBoost introduced a maximum prediction discrepancy of $9.35\times10^{-7}$, primarily associated with XGBoost's single-precision tree representation.

\textbf{Timing protocol.}
We measured the complete latency of each public explanation call rather than isolated CUDA-kernel time. Every timed call received a fresh contiguous host-array copy, and a fresh XGBoost \texttt{DMatrix} was constructed inside every Quadrature-TreeSHAP call; neither input containers nor previously computed explanations were cached. Model training, model conversion, explainer construction and initial CUDA compilation were excluded from the timed region. Every call was followed by explicit GPU synchronization, and numerical CPU libraries were restricted to one thread. For each seed we performed three warm-up calls followed by nine synchronized timed repetitions; the reported time for a model--batch cell is the median of the three seed-level medians and refers to the complete batch, not to an individual observation.

\begin{table}[H]
\centering\small
\caption{GPU tree benchmark: representative median runtime in milliseconds for a complete batch of size $B$ (NVIDIA RTX A6000; lower is better).}
\label{tab:tree_gpu_bench}
\begin{tabular}{@{}lrrrr@{}}
\toprule
Model & $B$ & \QuadraSHAP{} & Quadrature-TreeSHAP & GPUTreeSHAP \\
\midrule
1 tree, $d=16$, 256 leaves           & 1  & \textbf{0.410} & 1.327          & 2.407 \\
1 tree, $d=16$, 4{,}096 leaves       & 10 & \textbf{1.564} & 30.620         & 7.259 \\
1 tree, $d=16$, 16{,}384 leaves      & 10 & \textbf{1.952} & 121.789        & 22.307 \\
1 tree, $d=100$, 4{,}096 leaves      & 64 & \textbf{3.156} & 14.436         & 8.008 \\
32 trees, $d=16$, 128 leaves/tree    & 64 & \textbf{1.426} & 1.460          & 5.686 \\
32 trees, $d=100$, 128 leaves/tree   & 64 & 2.119          & \textbf{1.544} & 6.102 \\
\bottomrule
\end{tabular}
\end{table}

\textbf{GPU results.}
The exact, structure-dependent \QuadraSHAP{} implementation was faster than GPU Quadrature-TreeSHAP in 17 of the 18 model--batch cells. The geometric-mean speedup was $5.46\times$ and the largest speedup $62.38\times$, attained for the 16{,}384-leaf tree at batch size 10. The only Quadrature-TreeSHAP win occurred for the 32-tree, 100-feature forest at batch size 64, where it required $1.544$\,ms against $2.119$\,ms for \QuadraSHAP{}, a $1.37\times$ advantage; the corresponding 16-feature forest was approximately tied at this batch size. Against GPUTreeSHAP, \QuadraSHAP{} was faster in all 18 cells, with speedups from $2.48\times$ to $12.65\times$ and a geometric mean of $5.13\times$. The advantage increased notably for large single trees, where the quadrature formulation avoids the quadratic dependence on path depth of the original TreeSHAP recurrence.

\textbf{Equal-node control.}
Because the XGBoost implementation of Quadrature-TreeSHAP uses a fixed budget of eight quadrature evaluations, we conducted a separate control in which \QuadraSHAP{} was also forced to use $m_q=8$. Both methods used their native eight-point rules, so the number of function evaluations was matched although the transformed nodes and weights were not identical. This experiment used five warm-ups and 25 synchronized repetitions per seed. Under this equal-budget condition \QuadraSHAP{} remained faster in 16 of 18 cells, with a $5.03\times$ geometric-mean speedup and a maximum of $62.83\times$; Quadrature-TreeSHAP won the two batch-64 forest cases by $1.55\times$ and $1.59\times$. The forced-eight \QuadraSHAP{} output differed from its exact output by at most $4.10\times10^{-14}$ on this suite. The principal performance difference therefore cannot be explained by different quadrature-node counts.

\textbf{TreeGrad-Shap comparison.}
The official TreeGrad-Shap implementation was evaluated separately on the same six model shapes, three seeds and three batch sizes, with one warm-up and five timed repetitions. Since the released implementation is a single-row NumPy/Python CPU method, it ran on an AMD EPYC~9334 CPU, whereas \QuadraSHAP{} ran on the RTX A6000; for forests, TreeGrad-Shap was applied to each estimator separately and the attributions were averaged, as linearity permits. \QuadraSHAP{} was faster in all 18 cells, with speedups from $9.85\times$ to $3578.68\times$. These numbers compare the available implementations as complete systems and must not be read as an intrinsic algorithmic speedup, because devices, implementation languages and batching strategies differ substantially.

\textbf{Numerical agreement.}
In an independent, untimed validation, TreeGrad-Shap and exact \QuadraSHAP{} agreed to within $8.88\times10^{-15}$. The maximum attribution differences relative to exact \QuadraSHAP{} were $1.79\times10^{-6}$ for Quadrature-TreeSHAP and $7.26\times10^{-6}$ for GPUTreeSHAP, and exact \QuadraSHAP{} satisfied the efficiency axiom to within $1.42\times10^{-14}$. The larger discrepancies of the external GPU implementations are consistent with their single- or mixed-precision representations and are not evidence that they compute a different path-dependent game.

\textbf{Interpretation.}
The comparison with Quadrature-TreeSHAP does not establish a better asymptotic tree algorithm: both methods evaluate essentially the same quadrature representation. The observed difference is attributable to GPU execution and to the organization of the public call. \QuadraSHAP{} prepares its compact tree representation once, retains it on the GPU, evaluates subtree quantities in levelwise bottom-up and top-down passes, and reduces contributions grouped by feature without atomic updates. The tested XGBoost path reconstructs and transfers its compressed quadrature representation during each public explanation call, schedules work by row and tree with a depth-first traversal, and accumulates contributions with atomic additions. The specialized CUDA tree implementation does not use the generic product-game evaluator of \Cref{subsec:stable_efficient}, so the measured tree speedup should not be attributed to it. For GPUTreeSHAP, both methodology and implementation contribute: quadrature removes the quadratic path-depth factor, while the released GPUTreeSHAP wrapper also repeats path construction, transfer, allocation and bin-packing work for each explanation call. The TreeGrad-Shap result is strictly a comparison of available implementations, as no GPU implementation of that method exists. Finally, these experiments used one GPU, synthetic regression forests, depths of at most 24 and batches of at most 64, and they time end-to-end public explanation calls rather than isolated kernels. They support a strong low-latency and large-single-tree result on the tested system, not a claim of unconditional dominance across hardware, datasets or workload shapes.
\clearpage
\endgroup


\end{document}